\documentclass[10pt]{article}
\usepackage{amsmath}
\numberwithin{equation}{section}   
\usepackage{amssymb}
\usepackage{times}
\usepackage{graphicx}
\usepackage{color}
\usepackage{multirow}
\usepackage{url}
\usepackage{hyperref}
\usepackage{listings}
\usepackage{rotating}
\usepackage{bbm}
\usepackage{latexsym}

\newcommand{\captionfonts}{\normalsize}

\makeatletter  
\long\def\@makecaption#1#2{%
  \vskip\abovecaptionskip
  \sbox\@tempboxa{{\captionfonts #1: #2}}%
  \ifdim \wd\@tempboxa >\hsize
    {\captionfonts #1: #2\par}
  \else
    \hbox to\hsize{\hfil\box\@tempboxa\hfil}%
  \fi
  \vskip\belowcaptionskip}
\makeatother   

\definecolor{codegreen}{rgb}{0,0.6,0}
\definecolor{codegray}{rgb}{0.5,0.5,0.5}
\definecolor{codepurple}{rgb}{0.58,0,0.82}
\definecolor{backcolour}{rgb}{0.95,0.95,0.92}

\lstdefinestyle{mystyle}{
    backgroundcolor=\color{backcolour},   
    commentstyle=\color{codegreen},
    keywordstyle=\color{magenta},
    numberstyle=\tiny\color{codegray},
    stringstyle=\color{codepurple},
    basicstyle=\ttfamily\footnotesize,
    breakatwhitespace=false,         
    breaklines=true,                 
    captionpos=b,                    
    keepspaces=true,                 
    numbers=left,                    
    numbersep=5pt,                  
    showspaces=false,                
    showstringspaces=false,
    showtabs=false,                  
    tabsize=2
}

\newcommand{\eps}{{\epsilon}}

\newcommand{\pr}{\mathrm{pr}}

\newcommand{\cv}[1]{{\bf #1}}

\newcommand{\mat}[1]{{\bf #1}}
\newcommand{\setR}{{\mathbb{R}}}
\newcommand{\loss}{{\mathcal{L}}}

\newcommand{\tightitems}{\itemsep0pt\topsep0pt}

\begin{document}
\hspace{13.9cm}1

\ \vspace{20mm}\\

{\LARGE Supervised Hebbian learning in Deep Counterstream Associative Networks}

\ \\
{\bf \large Andreas Knoblauch$^{\displaystyle 1}$}\\
{$^{\displaystyle 1}$Albstadt-Sigmaringen University, KEIM Institute, Poststrasse 6, 72458 Albstadt-Ebingen, Germany, knoblauch@hs-albsig.de}\\
%


\thispagestyle{empty}
\markboth{}{NC instructions}
\ \vspace{-0mm}\\
%
\begin{center} {\bf Abstract} \end{center}
Modern machine learning applications employ deep neural networks training with the error backpropagation algorithm.
Although this algorithm is very effective, it lacks biological realism. For example, backpropagation requires symmetric connectivity,
and a separate neural processing channel for error signals. Prior works have therefore proposed a number of more realistic alternatives
for error backpropagation. However, most of them still suffer from demanding preassumptions that may be not fulfilled in the real brain,
for example, they often still require either symmetric connectivity or two separate processing channels, and often require also
special mathematical operations like subtractions or function inversions. Here I propose supervised counterstream learning in deep associative networks
as a simpler approach that requires only recognition of errors during training, and then backpropagates correcting target activity through
the same activity channel as used for forward propagation.
For this, two activity waves are initiated at the same time in input and output layers and then traveling in opposite directions to meet
in one of the hidden layers. By employing simple local Hebbian-type learning rules, the corresponding activity pattern sequences get linked bidirectionally,
thereby decreasing error rates over time. Despite its simplicity and an incomplete hyperparameter optimzation, a high high test accuracy is achieved
on the (binarized) MNIST data set that is comparable to more demanding architectures. 

\tableofcontents

\newpage

\section{Introduction}  \label{sec:introduction}

\begin{figure*}[ht]
  \begin{center}
  \includegraphics[width=\linewidth]{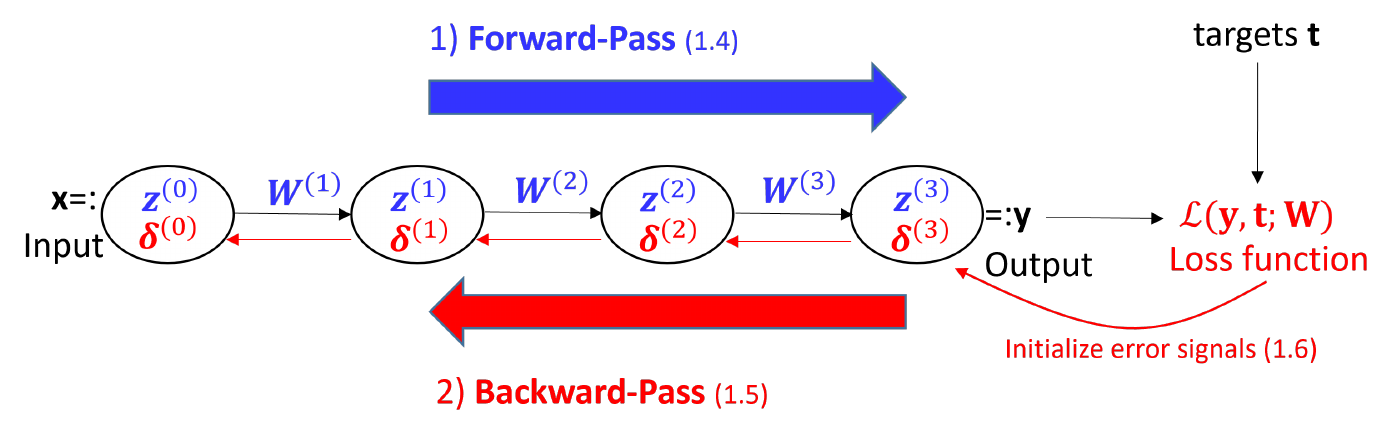}
  \end{center}
  \caption{\label{fig1:Backpropagation}
    Upating synaptic weights by backpropagation learning (\ref{eq:WeightUpdate_SGD},\ref{eq:partial_derivative_Wij}) requires a ``forward-pass''
    to compute firing rates $z^{(l)}_j$ of individual neurons $j$ in layer $l=1,\ldots,L$ and
    a separate ``backward-pass'' to compute error signals $\delta^{(l)}_j$.
  }
\end{figure*}

Artificial neural networks are chained (or multi-layered) model functions $\cv{y}(\cv{x};\mat{W})$ with model parameters (or synaptic weights)
$\mat{W}=(\cdots W_{ji}\cdots)$ that compute for input $\cv{x}=\cv{x}_n\in\setR^D$ a corresponding output $\cv{y}_n:=\cv{y}(\cv{x}_n;\mat{W})\in\setR^K$
(e.g., see \cite{Rumelhart/McClelland/PDPGroup:1986,Bishop:2006,Knoblauch:ILS:2026}).
Learning typically means minimizing a loss function
\footnote{
   Commonly used loss functions are, for example, sum of squared errors (SSE) with $\loss_n:=(y_{nk}-t_{nk})^2$ for $y_{nk},t_{nk}\in\setR$
   or categorical cross entropy (CCE) with $\loss_n:=-t_{nk}y_{nk}$ assuming one-hot-coding with $y_{nk}\in(0;1), t_{nk}\in\{0,1\}$ and $\sum_{k}t_{nk}=1$.  
}
\begin{align}
  \loss(\cv{W};\mathcal{D}) := \sum_{n=1}^N\loss_n(\mat{W};\cv{x}_n,\cv{t}_n) \quad\mbox{for}\quad \loss_n:=\sum_{k=1}^K\loss_{nk}(y_{nk},t_{nk}) =  \label{eq:lossL_WD}
\end{align}
with respect to the weights $\cv{W}$. Here the loss function represents the sum of component errors $\loss_{nk}(y_{nk},t_{nk})$
between network outputs $y_{nk}=y_k(\cv{x}_n;\mat{W})$ and corresponding
target values $t_{nk}$ given by the training data $\mathcal{D}=\{(\cv{x}_n,\cv{t}_n)|n=1,\ldots,N_{\mathrm{train}}\}$.
The minimization is often accomplished by a form of stochastic gradient descent \cite{Ruder:2016}
\begin{align}
  W_{ij}(\tau+1) = W_{ij}(\tau) + \Delta W_{ij}(\tau)  \quad\mbox{with weight change}\quad \Delta W_{ij}(\tau):=-\eta\frac{\partial\loss_n}{\partial W_{ij}}    \label{eq:WeightUpdate_SGD}
\end{align}
where $\eta>0$ is the learning rate and individual synaptic weights $W_{ij}$ from neuron $i$ to neuron $j$ get modified in proportion to the partial derivative of
the loss $\loss_n$ of the current input-output-pair $\cv{x}_n$ and $\cv{t}_n$ at time or learning step $\tau$. Gradient descent can be efficiently implement
applying the {\bf error backpropagation algorithm} \cite{Bryson/Ho:1969,Werbos:1974,Parker:1985,Rumelhart/McClelland/PDPGroup:1986}, where the partial derivates
\begin{align}
  \frac{\partial\loss_n}{\partial W_{ij}} = z_{ni}\delta_{nj}   \label{eq:partial_derivative_Wij}
\end{align}
are computed as the product of the presynaptic neuronal ``firing rate'' $z_{ni}$
and the postsynaptic ``error signal'' $\delta_{nj}$ for the current training input $\cv{x}_n$ (see Fig.~\ref{fig1:Backpropagation}).

%

The {\bf computational efficiency of backpropagation} comes from the fact that the synapse-related partial derivative (\ref{eq:partial_derivative_Wij})
decomposes into a simple product of neuron-related quantities $z_{ni}$ and $\delta_{nj}$ that can efficiently be computed for all neurons $j$ by a simple
forward- and backward-recursion through the network (e.g., see \cite{Bishop:2006}, chap.~5.3 or \cite{Knoblauch:ILS:2026}, chap.~5.2.1 for detailed derivations),
\begin{align}
  z_{nj}&:=h_j(a_{nj}) \quad\mbox{for}\quad a_{nj} = \sum_{i\in\mathrm{pre}(j)}W_{ij}z_{ni} \label{eq:backprop_forward_pass} \\
  \delta_{nj}&:=\frac{\partial\loss_n}{\partial a_{nj}}=h_j'(a_{nj})\cdot \alpha_{nj}  \quad\mbox{for}\quad \alpha_{nj} = \sum_{k\in\mathrm{succ}(j)}W_{jk}\delta_{nk} \label{eq:backprop_backward_pass}
\end{align}
where $a_{nj}$ is called the dendritic activation or dendritic potential, $h_j:\setR\rightarrow\setR$ is the activation function, $\mathrm{pre}(j)$ is the set of
predecessors of neuron $j$,
$h_j'$ is the derivative of $h_j$, $\alpha_{nj}$ is called the ``error potential'' of neuron $j$, and $\mathrm{succ}(j)$ is the set of successors of neuron $j$ in the network.
The recursions (\ref{eq:backprop_forward_pass}) and (\ref{eq:backprop_backward_pass}) are also referred to as {\bf forward-pass} and {\bf backward-pass}, respectively.
While the forward-pass is initialized with clamping the input layer neurons $z_{ni}=x_{ni}$ to the current input pattern $\cv{x}_n$ ($i=1,\ldots,D$),
the backward-pass is initialized with directly setting the error signals of output neurons $k$ to the partial derivatives of the loss function,
\footnote{
   In the three {\bf standard cases} of (1) linear output layer with SSE loss, (2) sigmoid outputs with binary cross entropy (BCE) loss, and (3) softmax outputs with CCE loss, 
   it follows $\delta_{nk}=y_{nk}-t_{nk}$, that is, error signals of output neurons $k$ are initialized with the ``difference error'' between network outputs $y_{nk}$
   and targets $t_{nk}$. For details see, e.g., \cite{Knoblauch:ILS:2026}, chap.~5.2.2. 
}
\begin{align}
   \delta_{nk}:=\frac{\partial\loss_n}{\partial a_{nk}} \quad (=y_{nk}-t_{nk}\quad \mbox{for standard cases}).  \label{eq:backprop_backward_init}
\end{align}
Note that each learning step in a network with $R$ synapses costs only $O(R)$ computing operations and is therefore optimal, whereas a naive
numerical evaluation of $\frac{\partial\loss_n}{\partial W_{ij}}$ costs $O(R^2)$ and is therefore infeasible in large networks.
\footnote{\label{footnote:EfficienceBackProp}
   Both forward pass and backward pass ``touch'' each of the $R$ synapses $W_{ij}$ exactly once, therefore computing costs are $O(R)$.
   Same holds for the synaptic updates (\ref{eq:WeightUpdate_SGD})
   with (\ref{eq:partial_derivative_Wij}) requiring a single multiplication per synapse. By contrast, a naive approximation of the partial derivatives by
   $\frac{\partial\loss_n}{\partial W_{ij}}\approx \frac{\loss_n(W_{ij}+\eps)-\loss_n(W_{ij}-\eps)}{2\eps}$ requires two forward-passes (\ref{eq:backprop_forward_pass})
   per synapse, resulting in $O(R^2)$. 
}

Although backpropagation is very efficient for training artificial neural networks in technical application, it is an {\bf unrealistic model of learning in the brain}
(for detailed discussions see \cite{Whittington/Bogacz:BackpropBrain:2019,Tang_etal:BioTraining:2022,Schmidgall_etal:braininspired:2024,Alkam_etal:RL_CognSc:2025}
or \cite{Knoblauch:ILS:2026}, chap.~8.6).
For example, forward and backward passes assume symmetric synaptic weights, as each synapse $W_{ij}$ that is used in the forward-pass (\ref{eq:backprop_forward_pass}) in forward-direction
(transmitting firing activity from neuron $i$ to neuron $j$) is also required in the backward-pass (\ref{eq:backprop_backward_pass}) in backward-direction
(transmitting error signals from neuron $j$ to neuron $i$).
By contrast, the neural networks in the brain are known to be quite asymmetric, both on a microscopic and macroscopic scale
\cite{Braitenberg/Schuz:1991,Felleman/VanEssen:1991}.
At least it has been shown that backpropagation works principally also with asymmetric weights using fixed random
backward connection, where, due to the so-called {\bf weight} or {\bf feedback alignment}'' phenomenon occuring in underdetermined networks,
the forward connections will adapt to be approximately symmetric, and by that achieve good performance comparable to symmetric
backpropagation \cite{Lillicrap_etal:FeedbackAlignment:2016,Song_etal:ConvergenceAlignment:2021}. Still, such approaches cannot easily explain
how real neurons should represent {\bf two different types of signals}, like activity signals $a_j, z_j$ versus error signals $\alpha_j, \delta_j$.

In {\bf this manuscript} I propose a novel approach to supervised learning that relies only on synaptic transmission of activity signals
(not requiring neuron-specific error signals $\delta_{nj}$)
and local Hebbian-type synaptic plasticity \cite{Hebb:1949,Steinbuch:1961,Willshaw/Buneman/Longuet-Higgins:1969,Palm:1982,Hopfield:1982,Lansner/Ekeberg:1989,Willshaw/Dayan:1990,Palm:1992,Palm/Sommer:1996,Lansner:2009,Knoblauch_BayesAsso:HRI2009,Knoblauch_BayesAssoIJCNN:2010,Knoblauch:NeurComp2011,Knoblauch_etal:Frontiers2012,Knoblauch:Elsevier2017,LansnerRavichandranKnoblauchHerman:arxiv:2025},
where synaptic changes $\Delta W_{ij}$ depend only on pre- and postsynaptic ``spike'' activity $z_i$ and $z_j$ modulated multiplicatively by
a global, dopamine-like error signal $\delta_n$ that expresses reward mismatches or the difference between network output $\cv{y}_n$ and targets $\cv{t}_n$.
The network is initialized as ``associative memory'' connecting local Hebbian cell assemblies \cite{Hebb:1949,Braitenberg:1978,Palm:1982,Knoblauch/Palm:2002_a,Knoblauch/Palm:2002_b,Pulvermuller:2003_a,Palm/Knoblauch/Hauser/Schuz:BiolCyb2014} forming activity attractors to foster invariant recognition.
As spikes are binary events, section~\ref{sec:binarization} describes methods for binarizing graded inputs like gray-value or RGB images. Section~\ref{sec:DCSAN} describes the network architecture
and the counterstream learning operations replacing backpropagation. Section~\ref{sec:experiments} reports results from numerical experiments implementing Deep Counterstream Associative Networks.
Finally, section~\ref{sec:discussion} discusses the implications of the results and possible future steps. 

\section{Binarization methods}  \label{sec:binarization}
\begin{figure*}[ht]
  \begin{center}
  \includegraphics[width=\linewidth]{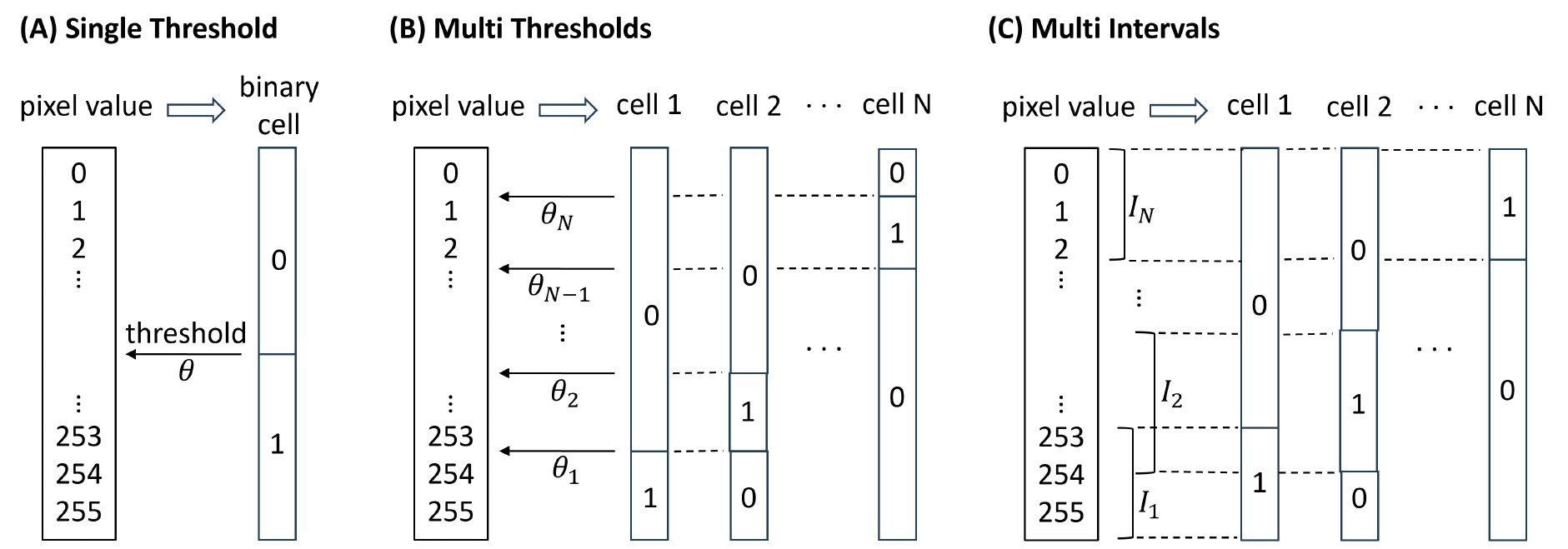}
  \end{center}
  \caption{\label{fig2:BinarizationImages}
    Threshold-based binarization methods to convert gray-scale or RGB images into binary pattern vectors. {\bf A}: The {\bf Single Threshold Method} applies for each pixel (of a channel)
    a threshold $theta$: Pixel values reaching (or exceeding) the threshold become 1, all others become 0.
    {\bf B}: The {\bf Multi Threshold Method} applies multiple thresholds $\theta_1>\theta_2>\cdots >\theta_N$ to activate multiple cells. Cell $j$ becomes 1 iff
    the pixel value lies in the interval $[\theta_j;\theta_{j-1})$ for and 0 otherwise (for $j=1,\ldots,N$ and defining $\theta_{0}:=\infty$).
    Note that for $\Theta_N:=-\infty$ this induces a $1$-of-$N$ block code, where each pixel is coded by a block of length $N$ having 1 active unit.
    {\bf C}: The {\bf Multi Interval Method} has also $N$ binary cells where cell $j$ becomes 1 iff the pixel value is in the interval $I_j$. Note that for overlapping
    intervals multiple units within a block (representing a pixel) can be activated at the same time. By choosing regularly overlapping intervals one may obtain a $K$-of-$N$ block code,
    where each block of length $N$ has $K$ active units.
  }
\end{figure*}

As many learning rules for associative networks assume binary ``spike'' activity \cite{Steinbuch:1961,Willshaw/Buneman/Longuet-Higgins:1969,Palm:1982,Hopfield:1982,Lansner/Ekeberg:1989,Willshaw/Dayan:1990,Palm:1992,Palm/Sommer:1996,Lansner:2009,Knoblauch_BayesAsso:HRI2009,Knoblauch_BayesAssoIJCNN:2010,Knoblauch:NeurComp2011,Knoblauch:Elsevier2017,LansnerRavichandranKnoblauchHerman:arxiv:2025},
we first have to transform gradual input vectors or images $\cv{x}_n\in\setR^D$
into corresponding binary input patterns $\cv{u}_n\in\{0,1\}^{D'}$. The simplest approach is to apply a {\bf single threshold} $\theta\in\setR$ to each input component
(see Fig.~\ref{fig2:BinarizationImages}A),
\begin{align}
  u_{ni}&:=\begin{cases} 1 &, x_{ni}\ge \theta \\
                        0 &, x_{ni}< \theta \end{cases}  \quad\quad\mbox{for $i=1,\ldots,D=D'$}.   \label{eq:single_threshold}
\end{align}
%

To reduce information loss of binarized compared to gradual inputs, we may similarly apply {\bf multiple thresholds}
$\theta_1>\theta_2>\cdots >\theta_N$ where (see Fig.~\ref{fig2:BinarizationImages}B)
\begin{align}
  u_{nij}&:=\begin{cases} 1 &, x_{ni}\in [\theta_j;\theta_{j-1}) \\
                         0 &, \mbox{else}\end{cases}  \quad\quad\mbox{for $i=1,\ldots,D$, $j=1,\ldots,N$, $\theta_0:=\infty$, $D':=DN$}.   \label{eq:multi_threshold}
\end{align}
Note that for $\Theta_N:=-\infty$ this induces a $1$-of-$N$ block code \cite{Wu:1982,Palm:1987_c,Kanter:1988,Kryzhanovsky/Litinskii/Mikaelian:2004,Lundqvist_etal:2006,Kryzhanovsky/Kryzhanovsky/Fonarev:2008,Kryzhanovsky/Kryzhanovsky:2008,Lansner:2009,Lundqvist_etal:2010,Gripon/Berrou:2011,Gripon/Rabbat:2013,Aliabadi/Berrou/Gripon/Jiang:2014,Aboudib/Gripon/Jiang:2014,Ferro/Gripon/Jiang:2016,KnoblauchPalm:ICANN2019,KnoblauchPalm:NeurComp2020,LansnerRavichandranKnoblauchHerman:arxiv:2025}, where each pixel is coded by a block of length $N$ having exactly 1 active unit.

A third binarization method is to apply {\bf multiple intervals} $I_j:=[\min_j;\max_j)$ for $j=1,\ldots,N$
with $\min_j,\max_j\in\setR\cup\{\pm\infty\}$, $\min_j<\max_j$ where (see Fig.~\ref{fig2:BinarizationImages}C) 
\begin{align}
  u_{nij}&:=\begin{cases} 1 &, x_{ni}\in I_j \\
                         0 &, \mbox{else}\end{cases}  \quad\quad\mbox{for $i=1,\ldots,D$, $j=1,\ldots,N$, $D':=DN$}.   \label{eq:multi_intervals}
\end{align}
By choosing the intervals in a regularly overlapping way, one may obtain a $K$-of-$N$ block code, where each pixel is coded by a block of length $N$ having exactly $K$ active unit.
\footnote{
For example, we have seen that from the intervals $I_j:=[\theta_j;\theta_{j-1})$ from (\ref{eq:multi_threshold}) with $N+1$ thresholds $\theta_0>\theta_1>\theta_2>\cdots >\theta_N$,
  $\Theta_0:=\infty$ and $\Theta_N:=-\infty$ we get a 1-of-$N$ block code, because these intervals provide a complete and disjunct cover over $\setR$. 
  If we add another complete disjunct cover with $N$ additional intervals $I_{j+(N+1)}:=[\theta_j+\Delta\theta;\theta_{j-1}+\Delta\theta$ for $j=1,\ldots,N$,
  where thresholds and thus intervals are shifted by $\Delta\theta\in\setR$ we get similarly a 2-of-$2N$ block code (because now each pixel value is covered by two intervals).
  If we similarly add further additional intervals $I_{j+2(N+1)}:=[\theta_j+2\Delta\theta;\theta_{j-1}+2\Delta\theta$ we get a $3$-of$3N$ block code, and so on.
  Thus for a $K$-of-$N$ block code we may add $K$ covers of $\setR$, where the $k$-th cover is provided by the intervals $I_{j+k(N+1)}:=[\theta_j+k\Delta\theta;\theta_{j-1}+k\Delta\theta$,
  where $k=0,1,\ldots,K-1$ and, for example, $\Delta\theta:=(\theta_{j-1}-\theta_{j})/K$ for equidistant intervals (or $[\theta_j;\theta_{j-1})$ being the shortes interval).
}
Typically one would choose sparse block codes, where $K$ is much smaller than $N$ (cf., \cite{Palm:1987,Kanerva:1988,Palm/Sommer:1992,Palm/Schwenker/Sommer:1994,Waydo/Kraskov/Quiroga/Fried/Koch:2006,Knoblauch/Palm/Sommer:NeurComp2010,Gripon/Berrou:2011,Palm:2013,SaCouto/Wichert:2020,KnoblauchPalm:NeurComp2020}). Then there are only a fraction $p:=K/N\ll 1$ active units per binary pattern vector.
As only active ``spikes'' have to be propagated through the network, this will also reduce computional costs in the forward pass by factor $p$, giving $O(pR)$
instead of $O(R)$. 

%
\begin{figure*}[ht]
  \begin{center}
  \includegraphics[width=\linewidth]{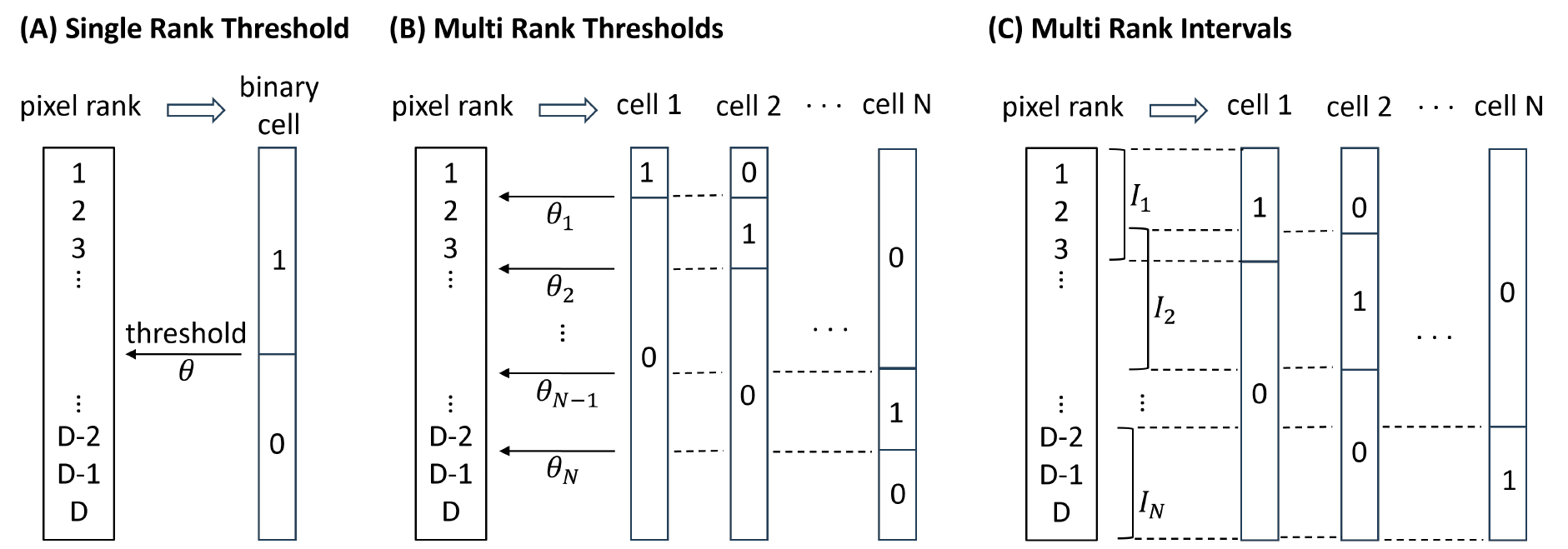}
  \end{center}
  \caption{\label{fig2B:BinarizationImagesRanks}
    Rank-based binarization methods to convert gray-scale or RGB images to binary pattern vectors. Similar to the previous Fig.~\ref{fig2:BinarizationImages},
    but thresholds and intervals are applied to pixel rank
    instead of pixel values.
  }
\end{figure*}

\begin{figure*}[ht]
  \begin{center}
  \includegraphics[width=\linewidth]{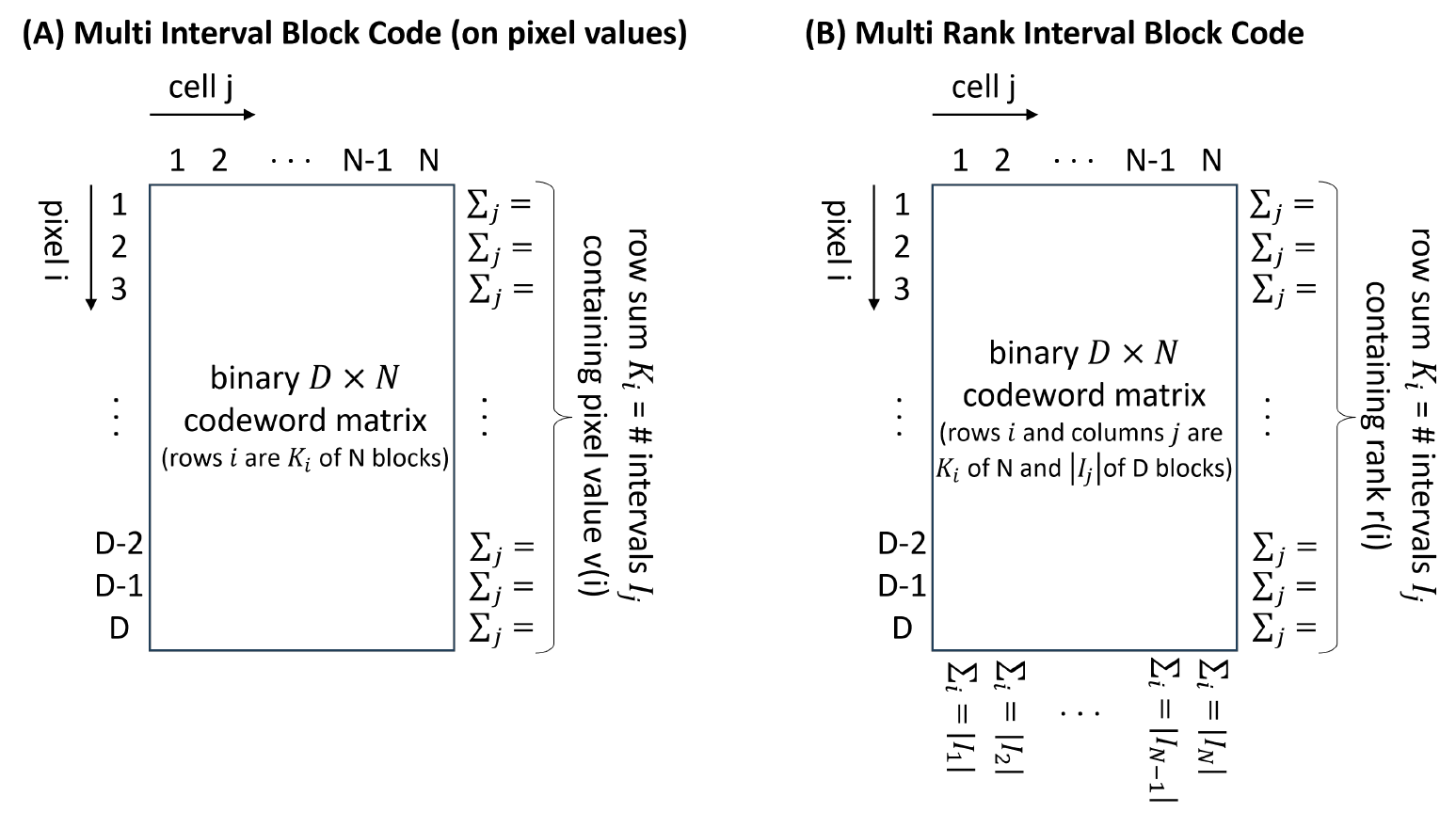}
  \end{center}
  \caption{\label{fig2C:BinaryCodewordMatrix}
    Binary codeword matrix for multi inverval block coding using either pixel values ({\bf A}) or pixel ranks ({\bf B})
    corresponding to previous Figs.~\ref{fig2:BinarizationImages},\ref{fig2B:BinarizationImagesRanks}.
    Note that the value-based binarization induces only a row-wise $K_i$-of-$N$ block structure, whereas the rank-based binarization
    induces also a column-wise $|I_j|$-of-$D$ block structure, where $|I_j|$ is the (integer) length of interval $I_j$.
    See text for details.
  }
\end{figure*}

Instead of applying thresholds on input values (or, in the case of input images, pixel values), the {\bf binarization} may also be {\bf based on the pixel value rank}: For $D$ dimensional
input vectors $\cv{x}_n\in\setR^D$, sorting the input components $x_{ni}$ for $i=1,\ldots,D$ in descending order gives as a ranking of the input indexes $i=1,\ldots,D$, that is,
$\mathrm{rank(i)}\in\{1,\ldots,D\}$ is the position of component $x_{ni}$ in the sorted input vector. As shown in Fig.~\ref{fig2B:BinarizationImagesRanks}, thresholds $\theta_j$ or intervals $I_j$
may then be applied to $rank(i)$, such that
(\ref{eq:single_threshold}-\ref{eq:multi_intervals}) become
\begin{align}
  u_{nij}&:=\begin{cases} 1 &, \mathrm{rank}(i)\in [\theta_j;\theta_{j-1}) \\
                         0 &, \mbox{else}\end{cases}  \quad\quad\mbox{for $i=1,\ldots,D$, $j=1,\ldots,N$, $\theta_0:=-\infty$, $D':=DN$}.   \label{eq:multi_rank_threshold}\\
  u_{nij}&:=\begin{cases} 1 &, \mathrm{rank}(i)\in I_j \\
                         0 &, \mbox{else}\end{cases}  \quad\quad\mbox{for $i=1,\ldots,D$, $j=1,\ldots,N$, $D':=DN$}.   \label{eq:multi_rank_intervals}
\end{align}
One {\bf advantage} of rank-based binarization is {\bf scale invariance}: The binarization $\cv{u}_n$ for input $\cv{x}_n$ equals the binarization of the scaled input $c\cv{x}_n$ for
any scaling factor $c>0$. Another advantage is that the resulting block patterns have a {\bf double block structure}: As illustrated by Fig.~\ref{fig2C:BinaryCodewordMatrix},
both value and rank based binarizations $\cv{u}_n\in\{0,1\}^{D\times N}$ consist of $D$ binary blocks of length $N$, each having a number of $K_i$ active units that corresponds
to the number of intervals $I_j$ that contain the pixel value $v(i):=x_{ni}$ (Fig.~\ref{fig2C:BinaryCodewordMatrix}A) or the rank number $\mathrm{rank}(i)$ (Fig.~\ref{fig2C:BinaryCodewordMatrix}A).
If each pixel value or rank number is covered by the same number $K_i=K$ of intervals, this results in a $K$-of-$N$ block codeword matrix $\cv{u}_n$, where each each row $i$ has the
same number $\sum_ju_{nij}=K$ of one-entries. By contrast, the column sums $\sum_iu_{nij}$ are undefined for the value based binarization (Fig.~\ref{fig2C:BinaryCodewordMatrix}A),
but correspond for rank-based binarizations simply to the interval lengths  $K'_j:=\sum_iu_{nij}=|I_j|:=\max(I_j\cap\{1,\ldots,D\})-\min(I_j\cap\{1,\ldots,D\})$
(Fig.~\ref{fig2C:BinaryCodewordMatrix}B). If all intervals $I_j$ have the same length $K'_j=K'$ this results in a $K'$-of-$D$ block code matrix. Thus, rank-based
binarizations can induce at the same time $K$-of-$N$ and $K'$-of-$D$ block structures.

%

This {\bf double block structure} could be {\bf exploited in technical applications} in two ways: First, it provides a more uniform distribution of binary patterns $\cv{u}_n=u^\mu$, where
the distribution of unit usages $M_1'(i):=\sum_\mu u_i^\mu$ of (\ref{eq:Muprime}) has smaller variance, which is well-known to benefit storage
in associative networks \cite{Buckingham/Willshaw:1992,Knoblauch:siam2008_WP}.
Second, it has been demonstrated that single block structure can significantly increase storage capacity of associative networks by allowing a $K$-of-$N$
winners-take-all retrieval strategy within each block (which maybe realized by local recurrent inhibition) or exploiting more subtle properties of the block structure \cite{Gripon/Berrou:2012,Yao/Gripon/Rabbat:2014,KnoblauchPalm:ICANN2019,KnoblauchPalm:NeurComp2020}.
It is, thus, likely that a double block structure will further enhance storage capacity, for example,
by implementing {\bf iterative retrieval} with {\bf alternating} row-wise $K$-of-$N$ and column-wise $K'$-of-$D$ {\bf winners-take-all selection}. While such retrieval procedures
may be advantageous in technical application, it may be {\bf more difficult} to {\bf interpret them neurobiologically}. Therefore most of the following experiments use simple
valued-based binarizations.

\section{Deep Counterstream Associative Networks}  \label{sec:DCSAN}
%
\begin{figure*}[ht]
  \begin{center}
  \includegraphics[width=\linewidth]{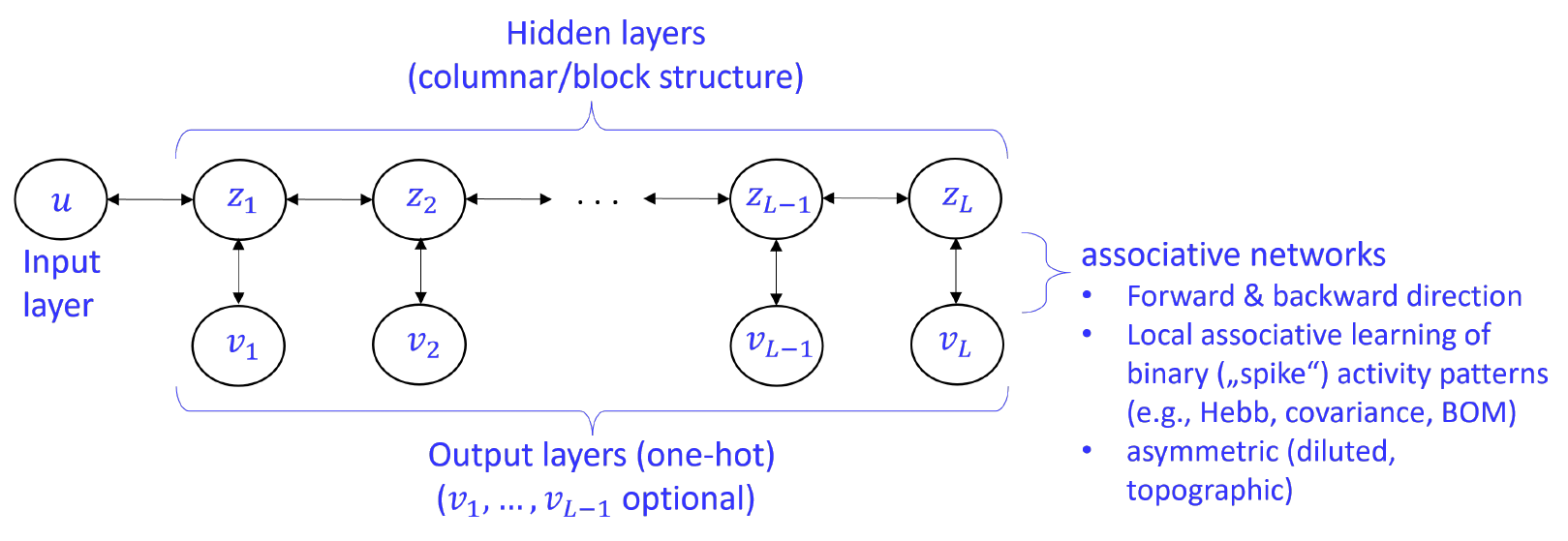}
  \end{center}
  \caption{\label{fig3:DeepCounterstreamNetwork}
    Deep Counterstream Associative Network: The network consists of an input layer $u$, $L$ hidden layers $z_l$, and $L$ output layers $v_l$ ($l=1,\ldots,L$).
    There are bidirectional associative synaptic connections between any two neighboring layers (double arrows mean two distinct
    associative connections in both directions). All layers have a columnar block structure (input layer $u$: 1 block per pixel value;
    hidden layers $z_l$: $B_l$ blocks with $K_l$ of $N_l$ active units; output layers: 1 block with 1-of-$C$
    active units (one-hot-coding). Synapses are trained using local associative learning rules, e.g., Steinbuch/Willshaw rule, Hebb rule, Hopfield rule, covariance rule or BOM rule \cite{Steinbuch:1961,Willshaw/Buneman/Longuet-Higgins:1969,Sejnowski:1977a,Palm:1980,Hopfield:1982,Willshaw/Dayan:1990,Knoblauch:NeurComp2011}.
    Associative networks are diluted (anatomical connectivity $P\le 1$), topographic (connection probability between two neurons depends on overlap of their receptive fields)
    and, in general, asymmetric. 
  }
\end{figure*}
For a {\bf biologically more realistic ``backpropagation'' learning} we have to get rid of error signals $\delta_{nj}$ symmetric networks $W_{ij}=W_{ji}$ as argued in the introduction. For this, let us briefly {\bf remember} the {\bf function of error signals} for learning: Negative error signals $-\delta_{nj}:=-\frac{\partial\loss_n}{\partial a_{nj}}$ provide simply information about how an increase of dendritic potential $a_{nj}$ will decrease the error $\loss_n$. Intuitively, it signals therefore to the synapes $w_{ji}$ connecting to neuron $j$ how effectful a weight modification could reduce the error by increasing or decreasing the postsynaptic dendritic potential $a_{nj}$. At the same time, this increase or decrease will be proportional to the presynaptic activity $z_{ni}$. therefore, for learning rate $\eta$, the {\bf backpropagation weight update rule} should indeed be
\begin{align}
   \Delta W_{ij} = \eta \cdot z_{ni}\cdot (-\delta_{nj})   \label{eq:learningrule_backprop_weightupdate}
\end{align}
as given by (\ref{eq:WeightUpdate_SGD},\ref{eq:partial_derivative_Wij}). For a {\bf more realistic Hebbian-type learning rule} we just have to replace the postsynaptic term $-\delta_{nj}$ by a spike-activity related quantity that also reflects the potential to decrease the error. {\bf Intuitively}, this quantity should be large for those postsynaptic neurons $z_j$ that have strong connections (over multiple layers) to the ``correct'' output neurons. As Hebbian-type learning rules tends to produce {\bf reciprocally connected cell assemblies}
\footnote{
  On the macroscopic level of layers, ``areas'' or ``macrocolumns'', cortex is well known to be symmetric:
  If there are connections from area $A$ to area $B$, then there are usually also reciprocal connections in reverse direction from area $B$
  to area $A$ \cite{Felleman/VanEssen:1991}. Thus, Hebbian-type learning rules (``what fires together wires together'') will tend to form cell assemblies with
  strong synaptic connections between all neurons that belong to the cell assembly. As a consequence, activating a part of the cell assembly in area $A$
  can activate the other part of the cell assembly in area $B$, and vice versa. 
  However, on the level of individual synpases, the connections are not (and need not be) strictly symmetric:
  First, the low anatomical connectivity $P\le 0.1$ within connected patches
  implies the same low probability that any synapse $W_{ij}$ from neuron $i$ to $j$ has a reciprocal partner synapse from $j$ to $i$.
  Second, even if a reciprocal partner exists, it will be unlikely that the weights $W_{ij}=W_{ji}$ would be equal, as required for symmetric connections.
  Third, it is actually known that the reciprocal feedback connections tend to be broader and less specific than the forward connections, and that they 
  tend to target different locations across cortical depth or dendritic arborizations. 
}
, these desired postsynaptic neurons could be activated by activating the ``correct'' target neurons $\cv{t}_n$ in the output layer (or perhaps just the
difference pattern $\cv{t}-\cv{y}$ between actual and predicted targets) and feeding back their activity through reciprocal
backward connections to the learning layer. This correponds to a kind of {\bf counterstream interaction} similar as suggested by \cite{Ullman:1995,Ullman:2021,Abel/Ullman:2024},
\footnote{\label{footnote:CounterstreamUllman}
   In the original counterstream model of Ullman et al. \cite{Ullman:1995,Ullman:2021} there is a bottom-up and top-down stream 
   integrating current inputs and expectations for {\it inference}.
   In newer works this idea is extended to combine inference with plausible learning mechanisms as an alternative to backpropagation \cite{Abel/Ullman:2024}.
   One major difference to the current model is that the counterstream of Ullman et al. is anatomically separated from the forward path, also requiring
   twin neurons for forward- and backward processing and ``counter-Hebbian learning''. By contrast, the DCAN proposed here makes fewer anatomical assumptions by
   using the same neurons for both forward- and backward processing.
}
where the {\bf forward stream} originating from the input layer provides the information about input-related activity $z_{ni}$ in the presynaptic layer, whereas
the {\bf backward stream} originating in the output layer provides the ``error-signal''related information about which postsynaptic neurons could
effecively activate the ``correct'' outputs and thus reduce the error.

One {\bf problem} of such counterstream interaction is that for typical applications like classification of objects on input images,
the feedback connections are {\bf not strictly reciprocal} in the mathematical sense, not even on the macro- or mesoscopic level of cell assemblies:
This is because many different input images converge to the same cell assemblies in the output layers. Therefore, activating a certain cell assemblies in
the output layer, feedback connections toward the input layer would spread activation potentially to an increasingly large number
of multiple cell assemblies per layer towards the input layer. Therefore it may be useful to have multiple output layers at each processing stage, in order
to improve invertibility of the cell assembly mappings between layers. Fig.~\ref{fig3:DeepCounterstreamNetwork} shows an instance of
a sequential {\bf Deep Counterstream Associative Network (DCAN)} consisting of an input layer $u$, followed by $L$ hidden layers $z_l$ for $l=1,\ldots,L$ and
an (obligatory) output layer $v_L$. Additionally, there may be (optional) output layers $v_l$ for any hidden layer $z_l$ ($l=1,\ldots,L-1$). 

\begin{figure*}[tb]
  \begin{center}
  \includegraphics[width=\linewidth]{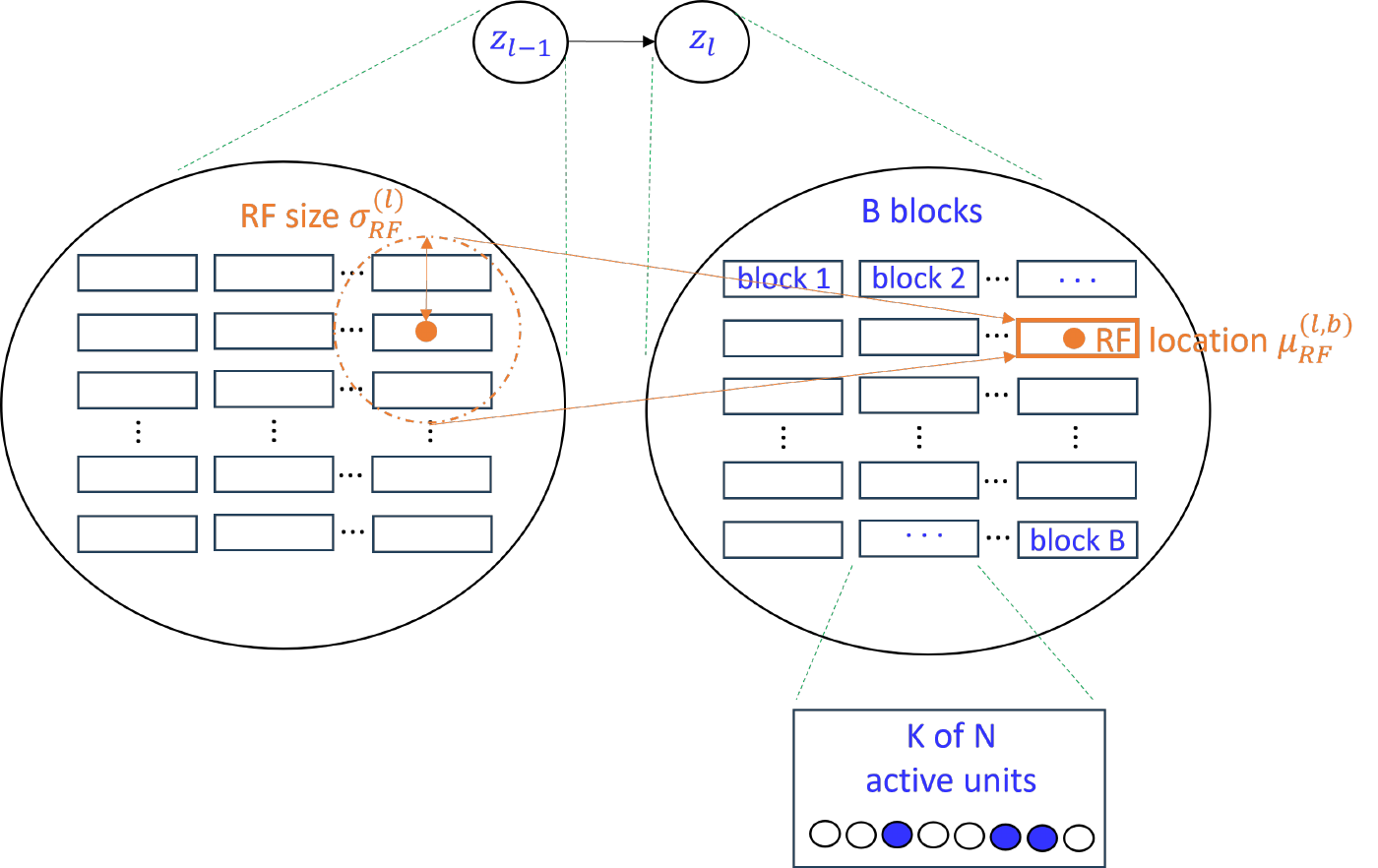}
  \end{center}
  \caption{\label{fig4:BlockStructureTopography}
    Block structure and topography of hidden layers: Each hidden layer $z_l$ consists of $B$ blocks (or synonymously modules or columns) of $N$ neurons,
    where during normal operation $K$ of the $N$ neurons per block get activated (by implementing a $K$-winners-take-all mechanism per block). 
    Each block $b_l$ of layer $z_l$ has a receptive field (RF) location $\mu_{\mathrm{RF}}^{(l,b)}$ (which is selected either regularly from a grid covering
    the input image area or randomly from a corresponding uniform distribution) and a RF ``size'' $\sigma_{\mathrm{RF}}^{(l,l-1)}$ (corresponding
    to the standard deviation of a Gaussian connection probability distribution; see text for details). 
  }
\end{figure*}

Layers are {\bf reciprocally connected} by forward- and backward-connections (bidirectional arrows) which are generally {\bf asymmetric} due to different connection probabilites, and different
topographic properties (backward connections having a larger variance; see below). Hidden-layers $z_l$ have a $K_l$-of-$N_l$ block structure, whereas output-layers employ $1$-of-$C$
``one-hot'' coding. For {\bf training} a local associative learning rule is used (like the Hebb rule, covariance or BOM rule, cf. app.~\ref{app:BOM},\ref{app:EWMA}).
Connections between hidden layers have a {\bf topographic}
organization as illustrated by Fig.~\ref{fig4:BlockStructureTopography}: Each block $b=1,\ldots,B$ of a layer $l$ has a {\bf receptive field (RF)} location $\mu^{(l,b)}\in[0;d_1]\times[0;d_2]$
where $d_1\times d_2$ is the size of the input layer or images. The ``chance'' of a connection between a neuron from block $b'$ of layer $z_{l-1}$ and a neuron from block $b$ of
layer $z_l$ is proportional to the Gaussian density $p_G\sim\exp(-(||\mu^{(l,b)}-\mu^{(l-1,b')}||/\sigma_{\mathrm{RF}}^{(l)})^2)+\mathrm{noise}$ with standard deviation $\sigma_{\mathrm{RF}}^{(l,l-1)}$.
More precisely, each neuron of layer $z_l$ receives $P^{(l,l-1)}n_{l-1}$ synapses from the neurons of layer $z_{l-1}$ with largest $p_G$ value, where we call $P^{(l,l-1)}$ the {\bf anatomical connectivity}
of the connection from layer $l-1$ to layer $l$ \cite{KnoblauchSommer:FNA2016,Knoblauch:Elsevier2017}. This results in topographic connections between layers
where neurons are densely connected if they (or rather their blocks) have similar RF locations, whereas they are only sparsely connected if their RF locations are further apart.

%

During {\bf training}, synaptic forward connections are first {\bf initialized} by defining random $K^{(l)}-of-N^{(l)}$ block patterns $z_l^\mu\in\{0,1\}^{n_l}$ for each hidden layer, clamping input and output layers to the
(binarized) training inputs $u^\mu$ and targest $v^\mu$, and then apply the local associative learning rule.
\footnote{
   This {\bf initialization} with random cell assemblies mimics ongoing Hebbian learning, for example, in prenatal developmental phases, expected to create Hebbian cell assemblies and columnar structure. If training inputs and outputs are
   presented for the first time, they are randomly linked to the pre-existing cell assemblies in the layers adjacent to input and output layers. 
}
The backward connections are initialized in a similar way, but using larger class-specific cell assemblies or activity patterns.
\footnote{
   The class-specific cell assemblies during initialization should be large enough such that each class assembly has sufficient overlap with each block in the hidden layer (at least on average).
}

%
\begin{figure*}[tb]
  \begin{center}
  \includegraphics[width=\linewidth]{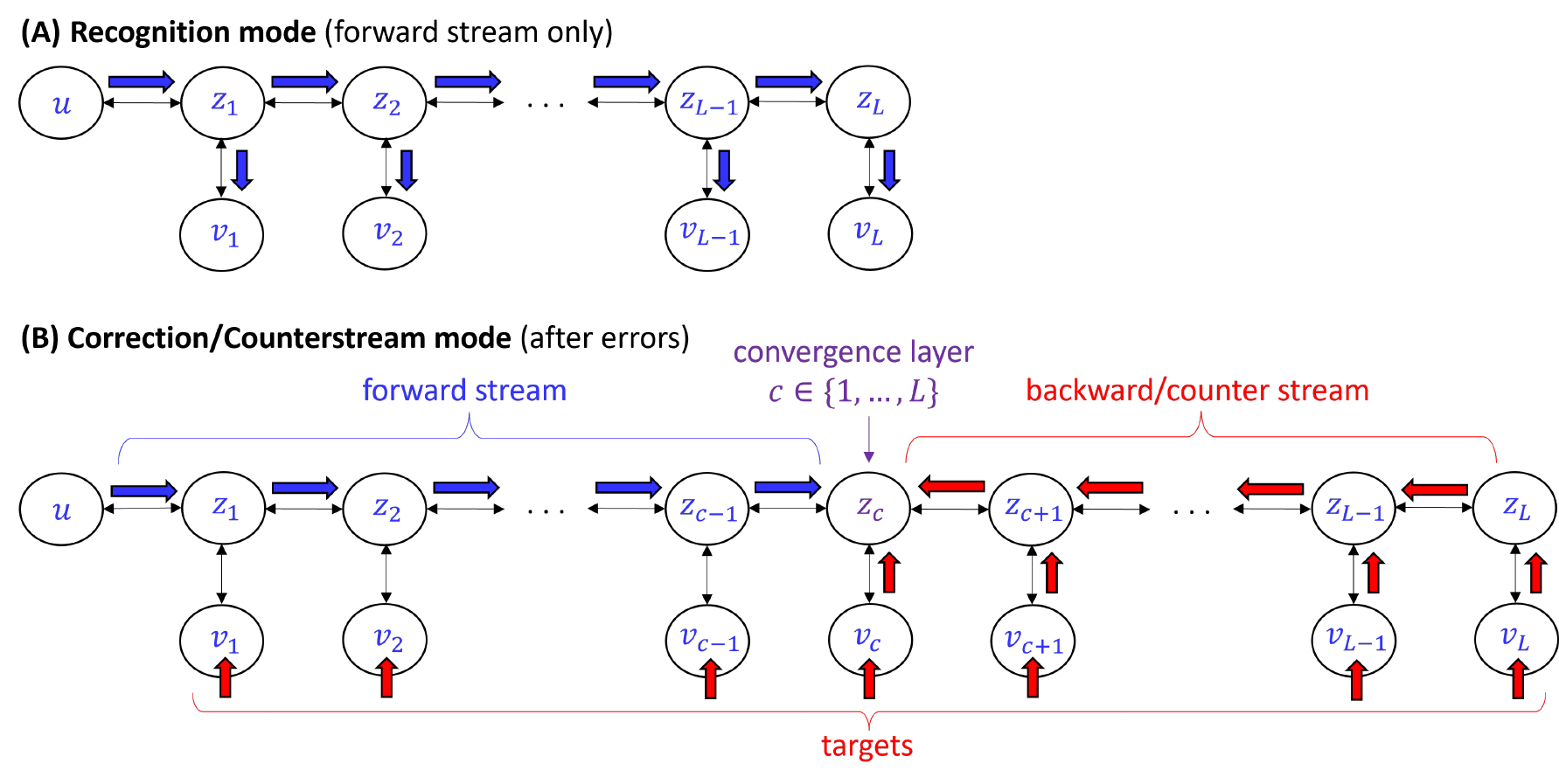}
  \end{center}
  \caption{\label{fig5:CounterStreamLearning}
    Operation modes of Deep Counterstream Associative Network. {\bf (A)} Recognition mode: An activity wave propagates in forward direction
    through the network (``forward stream''), transporting information about the input image to the hidden layers and
    producing activity vectors in the output vectors (typically representing certain classes of objects on the images to be recognized). 
    {\bf (B)} Correction or counterstream mode: During training, in particular after (A) has produced erroneous outputs,
    there may be a second activity wave traveling in backward direction from the output layers
    through the hidden layers towards the input layer (``backward or counter stream''), transporting information about the target values
    to the hidden layers. Forward and backward streams meet each other in the convergence layer $z_c$, where $c\in\{1,\ldots,L\}$ is selected
    either at random or following a fixed schedule. See text for details.
  }
\end{figure*}

After that, it follows {\bf ongoing continuous learning}
\footnote{
   The {\bf ongoing learning} is similar to a training phase in machine learning models. In brain models this may relate either to repeated presentation of sensory
   inputs and target signals (e.g., from a teacher saying
   the correct label or from reward related sources) or the replay from an episodic memory system
   like the hippocampus \cite{Wilson/McNaughton:1994,McClelland/McNaughton/OReilly:1995,Knoblauch_NCPW11:2009,Knoblauch/Korner/Korner/Sommer:PLOSONE2014,Knoblauch:Elsevier2017}.
}:
During an {\bf epoch}, training data is presented in random order. During each presentation, an input pattern is clamped to the input layer, and then propagated through the network in forward direction
(similar to the forward pass in backpropagation learning). In each block of each layer, the $K$ of $N$ most strongly activated neurons are activated by a $K$-winners-take-all mechanism. Similarly, in the output layers,
the most strongly activated ``one-hot'' neuron gets activated. 
If an input activates the {\bf correct outputs}
\footnote{
  As there may be multiple output layers, one has to define a criterion for ``correct trials'': For example, we may judge a trial as correct if the last output layer
  activated the correct one-hot neuron, or if all output layers had correct activities. We may also add a joint decision layer integrating (adding up) the inputs of all output layer (see below).
},
the activation patterns in the hidden layer are reinforced by the local associative learning rule, where weight updates are multiplied by a moderate factor $\delta_{\mathrm{corr}}$.
By contrast, {\bf erroneous outputs} will trigger {\bf counterstream learning} as illustrated in Fig.~\ref{fig5:CounterStreamLearning}, substituting the backward pass of backpropagation learning:
The forward stream corresponding to an activity wave starting in the input layer runs towards the output layers. Similarly, the backward stream corresponding to an activity wave starting in the output layers
runs towards the input layer. At some (predefined or randomly selected) layer the two waves converge and stop. The resulting activity patterns are then used again for Hebbian-type learning
applying the local associative learning rule, weighted by a typically larger factor $\delta_{\mathrm{err}}$.

\section{Explorative numerical experiments}\label{sec:experiments}

The purpose of the following experiments is to demonstrate the viability of supervised counterstream learning as described in the previous section as a realistic substitute of the backpropagation algorithm.
The {\bf basic setting} is training a deep associative network of {\bf $L=4$ hidden and output layers} with the (binarized) {\bf MNIST} dataset \cite{LeCun/Bottou/Bengio/Haffner:MNIST:1998,Deng:MNIST:2012} using the {\bf BOMs} learning rule (see eqs.~\ref{eq:bj_Bayesian_zeroerr_eps},\ref{eq:wij_Bayesian_zeroerr_eps}; cf. \cite{Knoblauch:NeurComp2011,LansnerRavichandranKnoblauchHerman:arxiv:2025}).
Besides synapses and biases there are a vast number of further hyperparameters that could be optimized. In informal preparatory experiments I used brief simulation runs to find
reasonable parameters by ``intuitive optimization'' as starting point for a more systematic hyperparameter optimization. Those {\bf initial hyperparameters} are:
$P=0.1$ (anatomical network connectivity); $B=50$ blocks per layer; $N=50$ neurons per block; $K=3$ active neuron per block (selected by winners-take-all);
random RF locations (distributed uniformly over image range); stride 1; RF sizes with $\sigma_{\mathrm{RF}}=4$ for feedforward and $\sigma_{\mathrm{RF}}=8$ for feedback connections
(connections to/from output layers were fully connected as they used one-hot-coding); dendritic noise for continous training was $\sigma_{\mathrm{corr}}=0.02$ in the initial phase,
$\sigma_{\mathrm{CSL}}=0.05$ during counterstream learning of erroneous trials, and $\sigma_{\mathrm{test}}=0$ during testing;
during a counterstream learning step of a minibatch, each layer was selected once to be the convergence layer; minibatch size 1000; 100 learning steps per experiments;
BOMs used numerical stability parameter $\eps=10^{-8}$ for feedforward and $\eps=10^{-30}$ for feedback connections; decay constant for moving averages of synaptic counters was $\beta=0.9$;
for reward-based reinforcement of local learning we used $\delta_{\mathrm{init}}=1$ for initial learning of ``class cell assemblies'', and for continuous training
$\delta_{\mathrm{corr}}=100$ for correct trials,
$\delta_{\mathrm{err}}=-100$ for erroneous trials, and $\delta_{\mathrm{CSL}}=100$ for counterstream learning; size of class asemblies was generally layer size divided by number of classes;
class assemblies were random patterns (without block structure) in hidden layers and one-hot-units in output layers; binarization of MNIST used single threshold with $\theta=150$; as input components of MNIST
only those pixels were preselected that were for at least 10 percent of training data $\ge\theta$ and for at most 90 percent of training data $<\theta$; this resulted in a binarization of MNIST data
with input layer size $n=292$ (instead of $28\times 28=784$) pixels with mean number of active units per input pattern $k=89.385$. In addition to the architecture shown in Fig.~\ref{fig3:DeepCounterstreamNetwork} there was an additional {\bf joint decision layer}
receiving (equally weighted) input from all output layers for computing final unique class decisions for each input. Both triggering counterstream learning (in case of errors) and evaluating test accuracy
was based on the output of the joint decision layer.

%
\begin{figure*}[tb]
  \begin{center}
  \includegraphics[width=0.75\linewidth]{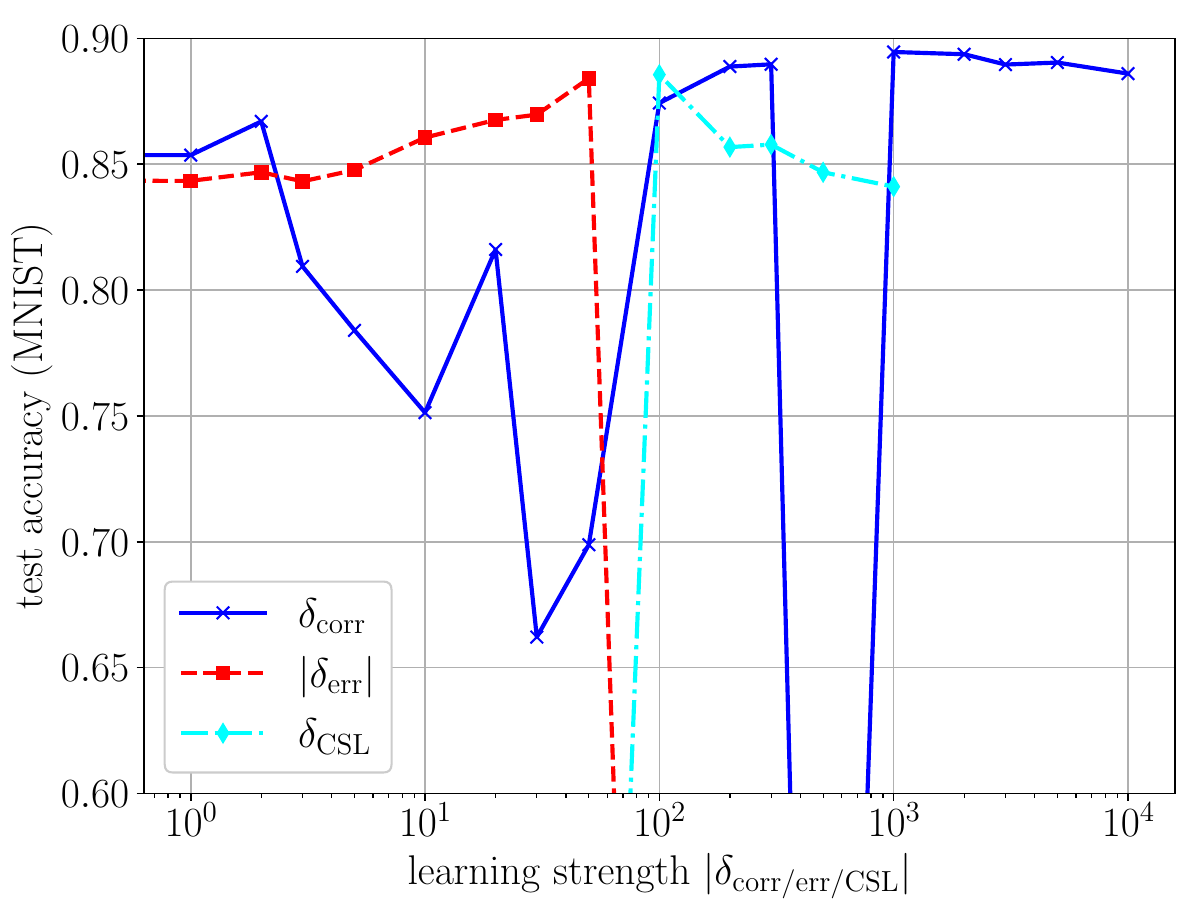}
  \end{center}
  \caption{\label{fig6:Eval1_DBOM_L_topo_acc_delta}
    Accuracy for MNIST test data as function of learning strength $\delta_{\mathrm{corr}}$ ({\bf Hebbian potentiation} after {\bf correct trials}; blue),
    $|\delta_{\mathrm{err}}|$ ({\bf Hebbian depression} after {\bf erroneous trials}; red), and
    $\delta_{\mathrm{CSL}}$ ({\bf Hebbian potentiation} after {\bf counter stream learning} following erroneous trials; cyan).
    Network had $L=4$ hidden layers $z_l$ and output layers $v_l$. Initial parameters were chosen reasonably
    according to a quick ad-hoc optimization for the experiments varying $\delta_{\mathrm{err}}$ (see text for details).
    Then optimal $\delta_{\mathrm{err}}=-50$ was chosen for all subsequent experiments doing ``{\bf coordinate ascent}'' optimization
    on MNIST test accuracy (max. accuracy was 0.8841). Similarly, subsequent experiments varying $\delta_{\mathrm{CSL}}$ yielded optimal $\delta_{\mathrm{CSL}}=100$ (with acc$=0.8856$),
    and subsequent experiments verying $\delta_{\mathrm{corr}}$ yielded optimal $\delta_{\mathrm{corr}}=1000$ (with acc$=0.8946$).
  }
\end{figure*}

After selecting initial hyperparameters, I used ``{\bf coordinate ascent}'' to maximize test accuracy. This means, starting from the initial hyperparmaters, each of the following experiments selects one (or a few) hyperparameters
to be optimized (and keeps the optimized parameters for all subsequent experiments, until they are optimized again). Fig.~\ref{fig6:Eval1_DBOM_L_topo_acc_delta} shows the results form {\bf experiments 1-3}, where
the three learning reinforcement factors $\delta_{\mathrm{err}},\delta_{\mathrm{CSL}},\delta_{\mathrm{corr}}$ were optimized sequentially (one after the other in the given order).
From {\bf experiment 1} we see that test accuracy increases from 0.84 to the maximum 0.88 with increasing
$|\delta_{\mathrm{err}}|$ (strength of Hebbian long-term-depression (LTD) after erroneous trials) from 0 to 50, but steeply breaks down above the optimum (red dashed line).
This shows that Hebbian LTD of erroneous trials can significantly improve performance, if it is not too strong. Next, {\bf experiment 2} used fixed $\delta_{\mathrm{err}}=-50$ and optimized  $\delta_{\mathrm{CSL}}$
(strength of Hebbian long-term-potentiation (LTP) during counterstream learning). It can be seen that test accuracy increases from 0.8411 to 0.8856 with decreasing $\delta_{\mathrm{CSL}}$ from 1000 downto 100,
but steeply breaks down below the optimum (cyan dash-dotted line). This shows that also counterstream learning with Hebbian LTP has an important role in learning. Finally, {\bf experiment 3}
used fixed previous optimal values $\delta_{\mathrm{err}}=-50$, $\delta_{\mathrm{CSL}}=100$ and optimized $\delta_{\mathrm{corr}}$ for Hebbian LTP of correct trials. Here maximal accuracy 0.8946 is reached for optimal 
$\delta_{\mathrm{corr}}=1000$ (blue solid line). Surprisingly there is at least one (or even two or three) further significant local maximum for smaller strengths.
One interpretation for this may be that the smaller local maximum comes from fine tuning LTP with respect to LTD, whereas the global maximum corresponds to a dominant ``correct trials'' LTP regime.
Therefore, {\bf experiment 4} re-optimized $\delta_{\mathrm{CSL}}$ (data not shown): The optimum $\delta_{\mathrm{CSL}}=100$ remained stable, but the drop for smaller values was shifted from $\delta_{\mathrm{CSL}}=50$ to $\delta_{\mathrm{CSL}}=10$,
indicating that the large $\delta_{\mathrm{corr}}=1000$ can compensate smaller $\delta_{\mathrm{CSL}}$. {\bf Together, experiments 1-4} confirm that all 3 learning mechanisms (LTP for correct trials, LTD and counterstream LTP for
erronous trials) are necessary and contribute to optimal performance. 

%
\begin{figure*}[tb]
  \begin{center}
  \includegraphics[width=0.75\linewidth]{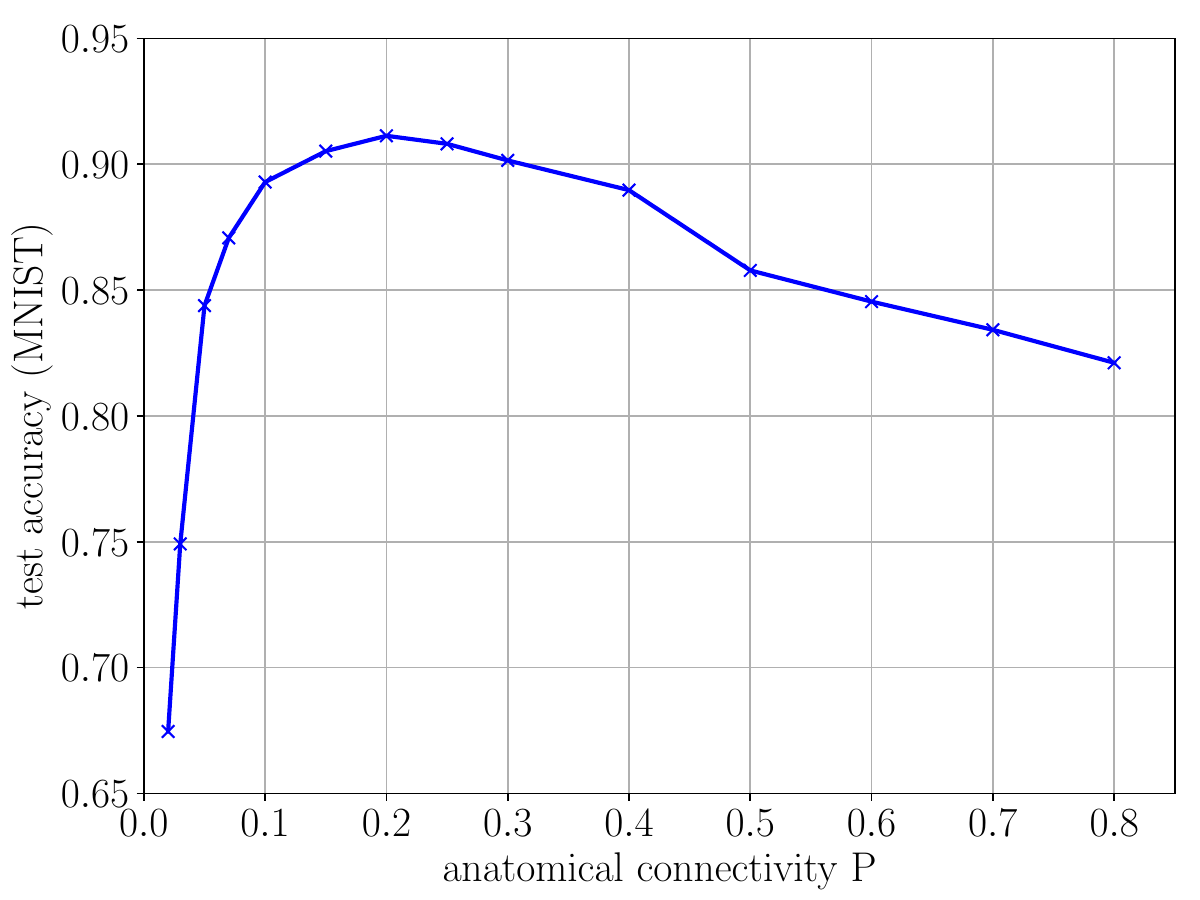}
  \end{center}
  \caption{\label{fig7:Eval2_DBOM_L_topo_acc_P}
    Accuracy for MNIST test data as function of anatomical connectivity $P$ of associative networks between input and hidden
    layers (associative networks to and from output layers were fully connected as they used a compact one-hot-code). Other parameters
    were chosen as in Fig.~\ref{fig6:Eval1_DBOM_L_topo_acc_delta}.
    Coordinate ascent was continued by selecting optimal $P=0.2$ (with acc$=0.9113$) for all subsequent experiments.
  }
\end{figure*}

Next, {\bf experiment 5} optimized {\bf anatomical network connectivity} $P$ in connections not involving output layers (that are fully connected, as mentioned above).
\footnote{
   Similar as before, previously optimized parameters keep their optimal value, i.e., $\delta_{\mathrm{err}}=-50$, $\delta_{\mathrm{CSL}}=100$, $\delta_{\mathrm{corr}}=1000$,
   thereby doing a ``coordinate ascent'' optimization. This holds also for
   all following experiments, although this is not mentioned anymore.
}
Fig.~\ref{fig7:Eval2_DBOM_L_topo_acc_P} shows results for $P$ between $2$ percent and $80$ percent (each neuron in the postsynaptic layer of a connection
receives exactly $Pm$ synapes from the presynaptic layer, where $m$ is the size of presynaptic layer). We see that there is a unique optimum at $P=0.2$ where
test accuracy reaches 0.9113. Actually, I would have expected from previous experiments with associative network that performance should increase monotonously
with anatomical connectivity $P$ (e.g., see \cite{KnoblauchSommer:FNA2016}, Fig.~7, althogh this corresponds to an associative memory task and not to classification).
On the one hand it is interesting that $P=0.2$ is close to local connectivity in the brain (which has been estimated at $P=0.1$ \cite{Braitenberg/Schuz:1991}, where performance is still good)
and that such low connectivity networks have efficient implementations also on digital hardware. However,
it may be that the optimum $P=0.2$ depends on the previously optimized learning strengths $\delta_{\mathrm{err}}=-50$, $\delta_{\mathrm{CSL}}=100$, $\delta_{\mathrm{corr}}=1000$,
and it may change for different values.
\footnote{
   In further experiments one should perhaps start with higher connectivity, e.g., $P=0.5$ or $P=0.8$ and then optimize all other parameters. I would predict that
   final test accuracy could further increase by such a procedure (note that our ``coordinate ascent'' procedure started with $P=0.1$ and yielded a higher optimum $P=0.2$).
   So it may be that starting with $P=0.5$, coordinate ascent will end with even higher $P>0.5$ (after optimizing learning strengths again). 
}

{\bf Experiment 6} optimized {\bf receptive field (RF) sizes} defined by standard deviation parameters $\sigma_{\mathrm{RF}}^{(l,l')}$ illustrated
in Fig.~\ref{fig4:BlockStructureTopography} (referring to the RF size of neurons from layer $l$ in presynaptic layer $l'$).
For this we tested different {\bf RF schemes} as given in the legends of Fig.~\ref{fig4:BlockStructureTopography}: {\bf Constant} means that all RF size parameters
$\sigma_{\mathrm{RF}}^{(l+1,l)}=\sigma_{\mathrm{RF}}^{(l,l-1)}$ have the same value. {\bf CNN-like} means similar to Convoluational Neural Networks used in machine learning, where
RF sizes $\sigma_{\mathrm{RF}}^{(l+1,l)}>\sigma_{\mathrm{RF}}^{(l,l-1)}$ get typically larger with layer number $l$. The forward$:$backward ratio defines the relative RF sizes
$\sigma_{\mathrm{RF}}^{(l+1,l)}:\sigma_{\mathrm{RF}}^{(l,l+1)}$ between forward and backward connections. It can be seen that smaller RF sizes in early layers
(compared to constant initial values $\sigma_{\mathrm{RF}}^{(l+1,l)}=4$ and $\sigma_{\mathrm{RF}}^{(l,l+1)}=8$) seem to perform better (CNN-like). Indeed, maximal test accuracy $0.9227$
was reached for CNN-like $\sigma_{\mathrm{RF}}^{(1,0)}=2, \sigma_{\mathrm{RF}}^{(2,1)}=4, \sigma_{\mathrm{RF}}^{(3,2)}=6, \sigma_{\mathrm{RF}}^{(4,3)}=8$ with forward$:$backward$=2:1$
(black dash-dotted).

%
\begin{figure*}[ht]
  \begin{center}
  \includegraphics[width=0.75\linewidth]{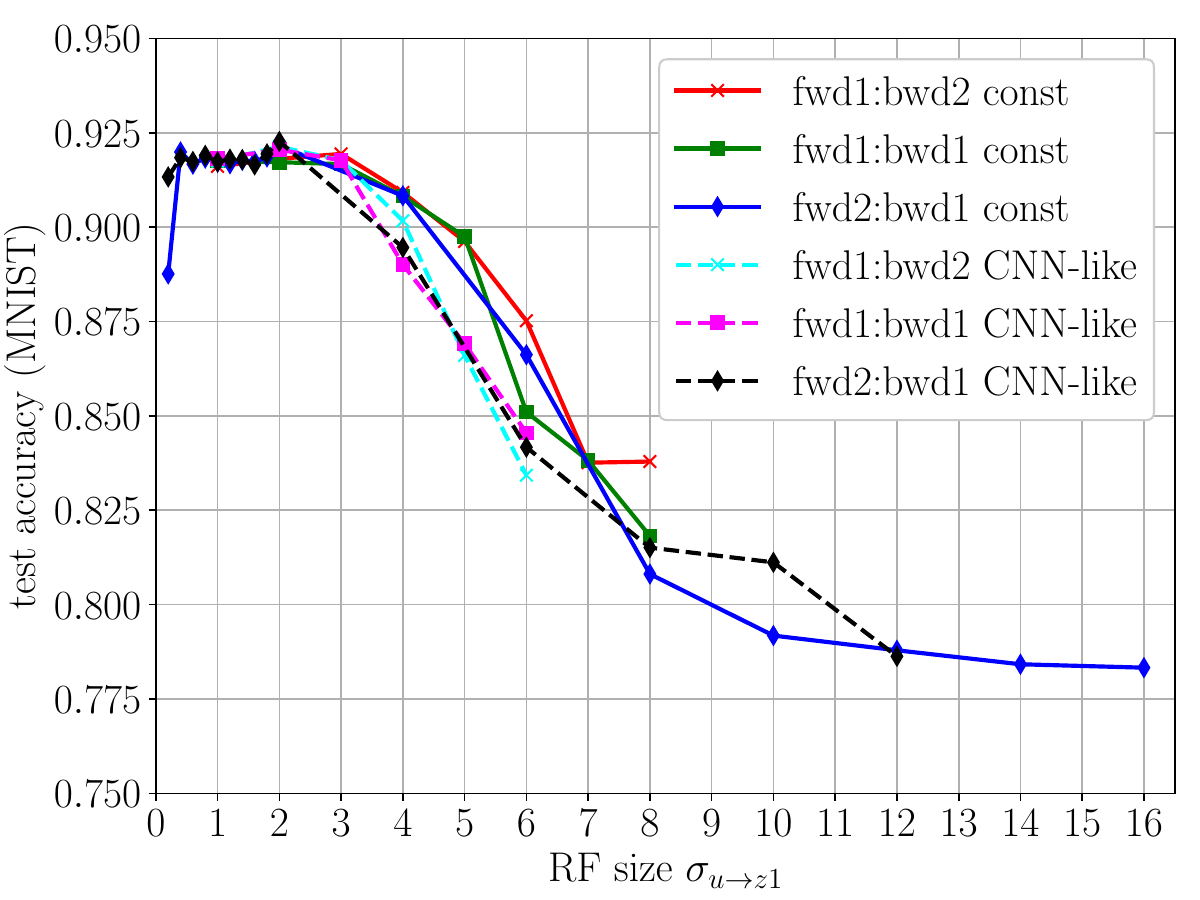}
  \end{center}
  \caption{\label{fig8:Eval3_DBOM_L_topo_acc_RFsize}
    Accuracy for MNIST test data as function of the receptive field (RF) size $\sigma_{u\rightarrow z1}$ of the associative network connecting input layer $u$ to first hidden layer $z_1$.
    Different RF schemes are used for the further associative networks: ``Const'' means that all networks in forward or backward direction used the same RF size
    (e.g., $\sigma_{u\rightarrow z1}$ for all forward networks), where ``fwdx:bwdy'' means that that the ratio of RF size in forward direction to backward direction was $x:y$.
    ``CNN-like'' means that RF size increases with layer index $l$. Other parameters
    were chosen as in Fig.~\ref{fig7:Eval2_DBOM_L_topo_acc_P}, continuing coordinate ascent. Maximum accuracy $0.9227$ was obtained
    for ``fwd2:bwd1'' with RF sizes $\sigma_{u\rightarrow z1}=2$, $\sigma_{z1\rightarrow z2}=4$, $\sigma_{z2\rightarrow z3}=6$, $\sigma_{z3\rightarrow z4}=8$ in forward direction
    (and half values $\sigma_{z2\rightarrow z1}=2$, $\sigma_{z3\rightarrow z2}=3$, $\sigma_{z4\rightarrow z3}=4$ in backward direction).
  }
\end{figure*}

\begin{figure*}[ht]
  \begin{center}
  \includegraphics[width=0.75\linewidth]{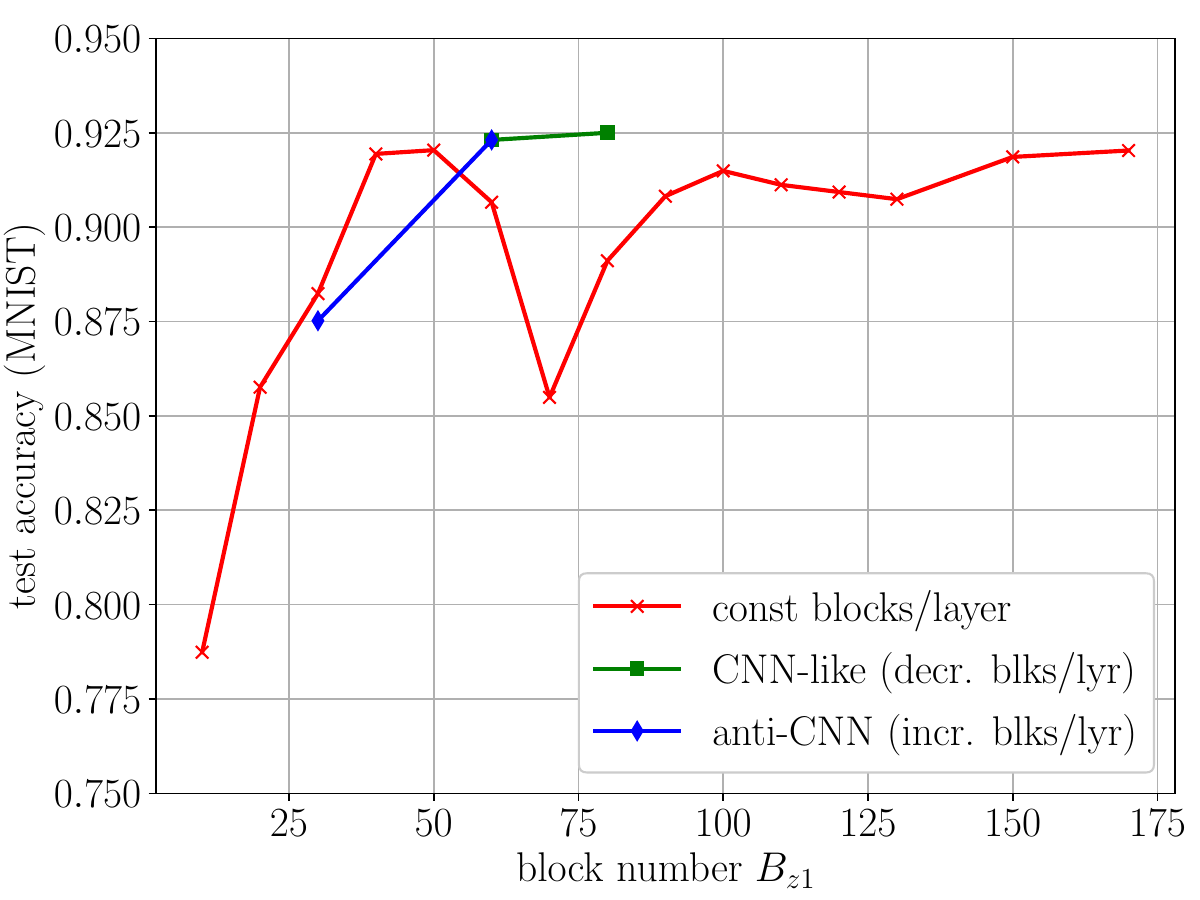}
  \end{center}
  \caption{\label{fig9:Eval4_DBOM_L_topo_acc_B}
    Accuracy for MNIST test data as function of block number $B_{z1}$ of layer $z_1$ (each block consisting of $N=50$ neurons of which $K=3$ neurons are active).
    Different block number schemes are used for the further layers: ``Const'' means that all layers have the same block number $B_{zl}=B_{z1}$
    for all futher layers $l=2,3,\ldots,L:=4$. ``CNN-like'' means that block number (corresponding to resolution) decreases with layer number $l=2,3,4,\ldots$.
    ``Anti-CNN'' means that block number increases with layer number $l=2,3,4,\ldots$. Other parameters
    were chosen as in Fig.~\ref{fig8:Eval3_DBOM_L_topo_acc_RFsize}, continuing coordinate ascent. Maximum accuracy $0.925$ was obtained
    for ``CNN-like'' with block numbers $B_{z1}=80$, $B_{z2}=50$, $B_{z3}=30$, $B_{z4}=20$. 
  }
\end{figure*}

{\bf Experiment 7} optimized {\bf block numbers} $B_l$ in hidden layer $l$ and, thus, implicitely also the {\bf layer sizes} $n_l=B_lN_l$ (where here we kept constant $N_l=50$ as
in the initial parameter set). Similar to RFs, we applied different {\bf block number schemes}: {\bf Constant} means that all layers have the same block number $B_l=B_{l+1}$.
{\bf CNN-like} means that block number $B_l>B_{l+1}$ decreases with layer index $l$ (similar to CNNs, where pixel resolution decreases with $l$). Conversely, {\bf anti-CNN} means
that block number $B_l<B_{l+1}$ increases with layer index $l$. As illustrated by Fig.~\ref{fig9:Eval4_DBOM_L_topo_acc_B}, for constant block number there are at least two local optima:
One is at $B_l=50$ and may result from the previous coordinate ascent optimization procedure (as it corresponds to initial value, used so far). Then there seems another local optimum
at $B_l=100$ and possibly a monotonous increase for large $B_l\rightarrow\infty$ (which may correspond to the well known fact that capacity of associative networks increases
with network size \cite{Tsodyks/Feigelman:1988,Knoblauch:NeurComp2011}).
However, the global maximum occurred for the CNN-like scheme with $B_1=80$, $B_2=50$, $B_3=30$, $B_4=20$ blocks, reaching test accuracy 0.925.
It should be noted that I have tested here only 4 non-constant configurations (2 CNN-like and 2 anti-CNN with relatively small block numbers),
and there will certainly be potential for further optimizations, in particular in the large-network regime with larger block numbers.

%

{\bf Experiment 8} did a joint optimization of {\bf block numbers} $B_l$ and {\bf connectivity} $P_l$ (data not shown).
I tested further configurations of CNN-like block numbers $B_l>B_{l+1}$ (with $B_1$ ranging between 30 and 100)
combined with different schemes for ``CNN-like'' layer-dependent connectivity $P_l<P_{l+1}$, where connectivity becomes denser for later/smaller layers. Connectivity from $u$ to $z_1$
was always $P_1=0.1$ und $P_4$ reached up to 0.8. Forward-to-backward ratio was 2:1, that connectivity of forward connection $z_l\rightarrow z_{l+1}$ hat double connectivity than corresponding
backward connection $z_l\leftarrow z_{l+1}$. Unfortunately, performance was quite bad, maximal test accuracy was only 0.9068, and so no improvement was reached. Similar to experiments 9
I have tested only a very limited set of possible configurations (still restricted to relatively small block numbers). Also, similar as argued for experiment 5, it may be that
learning strengths are suboptimal for higher connectivities. So, there might be still potential for improvements. 

{\bf Experiment 9} extended experiment 8 by doing a joint optimization of {\bf block numbers} $B_l$, {\bf connectivity} $P_l$, and {\bf RF sizes} (data not shown). Again, no improvement of test accuracy was
reached (max. test accuracy was 0.9191), but due to my limited evaluation of the search space, there may still be potential for improvements.

\begin{figure*}[ht]
  \begin{center}
  \includegraphics[width=0.75\linewidth]{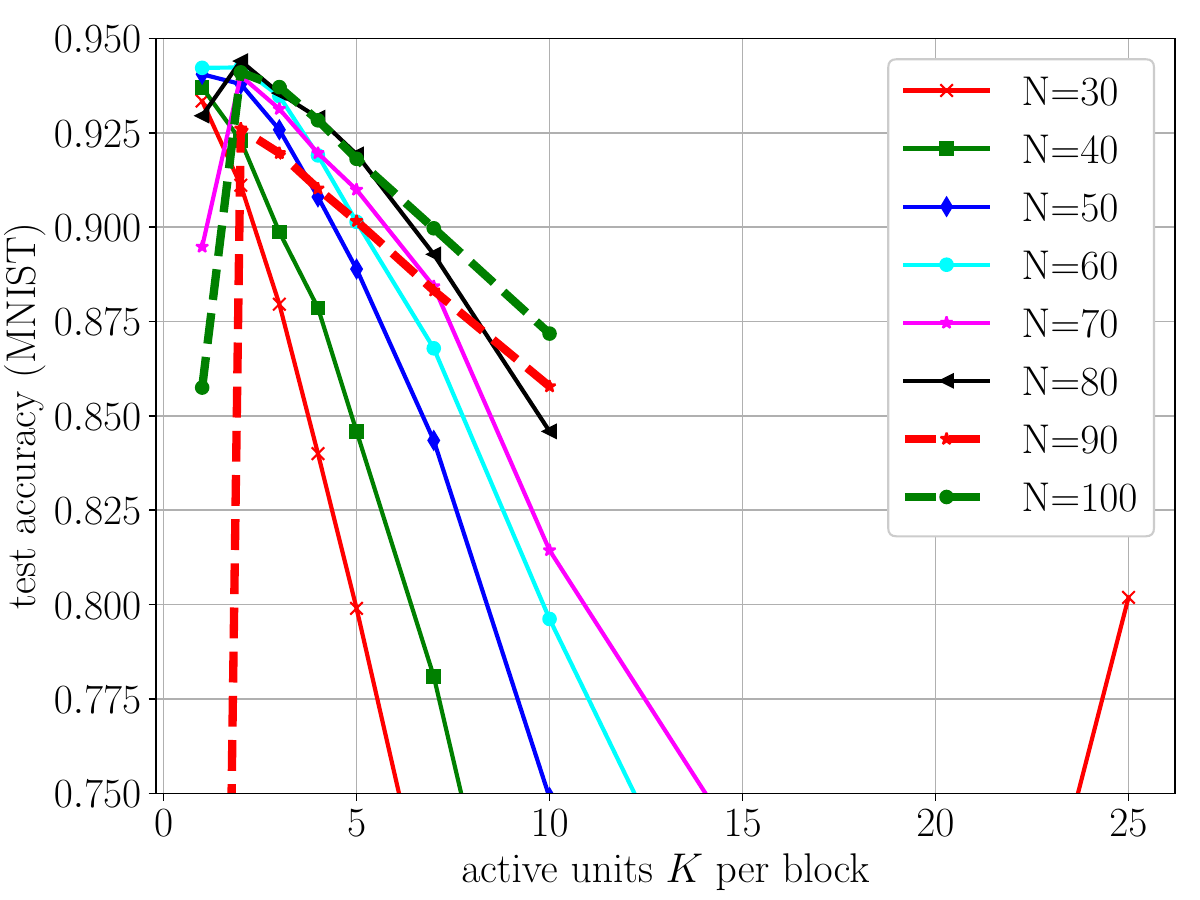}
  \end{center}
  \caption{\label{fig10:Eval5_DBOM_L_topo_acc_KN}
    Accuracy for MNIST test data as function of number of active neurons $K$ per block, using different block sizes $N$.
    Other parameters
    were chosen as in Fig.~\ref{fig9:Eval4_DBOM_L_topo_acc_B}, continuing coordinate ascent. Maximum accuracy $0.944$ was obtained
    for $K=2$ and $N=80$. Note that maximal accuracy requires very sparse coding with only one or two active units per block!
  }
\end{figure*}

Finally, {\bf experiment 10} did a joint optimization of $K$ ({\bf active units per block}) and $N$ ({\bf neurons per block}), always choosing constant parameter values over all layers, $K_l=K_{l+1}$ and $N_l=N_{l+1}$.
Fig.~\ref{fig10:Eval5_DBOM_L_topo_acc_KN} shows test accuracy as function of $K\in\{1,2,3,4,5,7,10,15,20,25\}$ for different $N\in\{30,40,\ldots,100\}$.
For any $N$ best results are obatined for very small $K\in\{1,2,3\}$, that is, high performance requires very sparse coding of neuronal activity within blocks. Performance falls steeply for larger $K$.
For small $N\le 50$, optimal $K$ was 1 corresponding to a standard $1-of-N$ block code as used in many previous works on associative networks \cite{Wu:1982,Palm:1987_c,Kanter:1988,Kryzhanovsky/Litinskii/Mikaelian:2004,Lundqvist_etal:2006,Kryzhanovsky/Kryzhanovsky/Fonarev:2008,Kryzhanovsky/Kryzhanovsky:2008,Lansner:2009,Lundqvist_etal:2010,Gripon/Berrou:2011,Gripon/Rabbat:2013,Aliabadi/Berrou/Gripon/Jiang:2014,Aboudib/Gripon/Jiang:2014,Ferro/Gripon/Jiang:2016,KnoblauchPalm:ICANN2019,KnoblauchPalm:NeurComp2020,LansnerRavichandranKnoblauchHerman:arxiv:2025}. 
By contrast, for larger $N>50$, optimal $K$ was 2 corresponding to a distributed $2-of-N$ block code. Maximum test-accuracy 0.944 was reached for $N=80$, $K=2$. 
I have done also some preliminary experiments showing that optimal $K>2$ can be larger for larger $N>100$ corresponding to a $K-of-N$ block code for still small $2<K\ll N$ (data not shown),
but test accuracy decreased further for $N>80$.

%

\section{Summary and discussion}\label{sec:discussion}

This report proposes supervised Hebbian-type learning in {\bf deep counterstream associative networks (DCAN)} as a biological alternative to
gradient descent through backpropagation learning employed in machine learning applications. DCAN requires only local Hebbian-type learning rules
(that depend on pre- and postsynaptic coincidences) and binary neuronal ``spike'' activity. Unlike error backpropagation \cite{Werbos:1974,Rumelhart/Hinton/Williams:1986,Bishop:2006},
it does require neither
symmetric connectivity nor a separate activity channel for backpropagating ``error signals'' corresponding to the partial derivatives of the loss function. Instead, for
erroneous training trials, counterstream learning is triggered, where activating target assemblies in the output layers will backpropagate
an activitity wave, that clashes with the feedforward wave originating from the input layer, thereby connecting bottom-up with
top-down information by local associative learning of spike activity only (cf. \cite{Ullman:1995,Ullman:2021,Abel/Ullman:2024}; see also footnote~\ref{footnote:CounterstreamUllman}
in section~\ref{sec:DCSAN}).  
Such counterstream learning seems to connect the forward wave
with the ``correct'' assemblies activated by the backward wave representing the true class label, increasing the chance that the same input
will in future trials activate the correct class representations. In addition to Hebbian long-term potentation (LTP), the model includes also
a mechanism for long-term-depression (LTD), where connections between activity patterns of the forward wave leading to a wrong classification will
be weakened. The switch between LTP and LTD is implemented by a multiplicative reward-based factor $\delta$ that modulates Hebbian
learning strength, being positive for LTP and negative for LTD. The reward factor $\delta$ may correspond to a dopamine-related
reward signal (see app.~\ref{app:EWMA}), as employed in reinforcement learning models \cite{Schultz/Dayan/Montague:1997,Schultz:1998,Fremaux/Gerstner:2016}.
Although dopamine release can only be positive in
the real brain, an effectively negative interaction could be emulated by interaction of a positive release with a continuous decay of synaptic
(coincidence) variables, for example, by a simple first-order moving average accumulation mechanism (see app.~\ref{app:EWMA}; cf.~\cite{Ravichandran/Lansner/Herman:FrontNeurosci:SpikingReprLearn:2024,Kingma/Ba:2015})    
Despite using only biologically plausible mechanisms and having done only a preliminary hyperparameter optimization,
DCAN can reach real-world classification results, for example, on the MNIST data set, that are comparable to current machine learning models,
and, if scaled-up to larger and more complex architectures, may be comparable to human performance.

There are a number of further biological alternatives to error backpropagation that have been propsed previously \cite{Whittington/Bogacz:BackpropBrain:2019,Lillicrap_etal:2020}. Some of them may reach
a few percent more performance than the DCAN model proposed here, but they make significantly stronger assumptions about the anatomy and the processing capabilities of the brain: For example,
typical {\bf weight/feedback alignment models} allow asymmetric connectivity consistent with real brain networks, but still require a second activity channel for backpropagation
of error signals \cite{Lillicrap_etal:FeedbackAlignment:2016,Nokland2016}, which is difficult to reconcile with known anatomical features of the brain. By contrast, DCAN does not requires an explicit channel for error signals.
Similarly, in {\bf predictive coding models} \cite{Whittington2017,Rao/Ballard:1999} the aim of each layer is to predict activity of the previous layer.
As learning is based on the prediction error, this
requires a sufficiently precise mirroring and subtraction between two separate activity channels (actual activity and predicted activity) and, thus, a very dedicated
microcircuitry for implementation, which is uncertain to exist. Moreover, many predictive coding models also require symmetric connectivity in forward- and backward direction.
By contrast, DCAN just requires bidirectional (not symmetric) associative connections between layers which
are well known to exist in the brain and, perhaps, some mechanisms to recognize errors in the output layers (without the need for a precise subtraction) to trigger counterstream learning.
Then {\bf target propagation models} may be seen similar in spirit to DCAN, but they employ approximate inverse mappings for an explicit computation of targets in each layer \cite{Lee_etal:DTP:2015,Ororbia_etal:LRAE:2019}.
Again, representing the targets requires a second mirroring activity channel and a comparison of feedforward activity with the target activity at each stage. By contrast DCAN just requires
backpropagating activity patterns through feedback connections that activate the same cell assemblies as during forward processing. These cell assemblies activated by the counterstream
may be interpreted as target signals, having strong forward-connections to the ``correct'' output neurons. Note that this assumes only a ``quasi-symmetric'' connectivity on the level
of cell assemblies, but not precisely symmetric connections on the level of individual neurons.
Another model that makes very few assumptions on brain anatomy and physiology is {\bf weight perturbation} or,
to a lesser degree, {\bf node perturbation} \cite{Zuege/Klos/Memmesheimer:WeightPerturb:2023,Jabri/Flower:WeightPerturbation:1992,Cauwenberghs:WeightPerturb:1993,StrickerRoehrbeinKnoblauch_IJCNN:2024,Williams:1992,Fiete_eta:NodePert:2007}.
Here the basic idea is to emulate gradient descent on the loss function by making random perturbations of either synaptic strength or neuron activity
(the latter being significantly less plausible). These changes are either reinforced if the error has decreased after perturbation, or inverted if the error has increased.
These models basically apply a random local search improved by reward signals and suffer from the combinatorial explosion of the search space in larger networks.
Moreover, they need a precise measurement (and again subtraction) of the loss function to avoid excessive perturbation noise effects.
Finally, there are unsupervised {\bf local Oja-like feedforward PCA-learning} networks that achieve surprisingly high performance if adding additionally
simple linear classification layers on top \cite{Oja:1982,Oja1992,Krotov/Hopfield:2019,Zhou:arxiv:2022,Inoue/Rohrbein/Knoblauch:GuidingSparseNN:2026,StrickerRoehrbeinKnoblauch:bwhpc:2026}. 
However, they also require some symmetry within local connections. Although this symmetry constraint may be relaxed \cite{Friston_etal:NeuroPCA:1993}, they seem to have problems or need to be
extended for supervised learning tasks.

Thus, although the DCAN model may not achieve the best performance among the biological alternatives of backpropagation, it may constitute one of the best models
balancing performance and biological realism. Also, the first explorative experiments shown in this report have likely not reached the optimal performance, due to an
incomplete/preliminary hyperparameter optimization.
In future work, it may be interesting to do a more comprehensive hyperparameter optimization to come closer to the performance limits of DCAN.
Also, the experiments should be extended from MNIST to more challenging data sets. Currently, only the BOMs learning rule has been tested
(see eqs.~\ref{eq:bj_Bayesian_zeroerr_eps},\ref{eq:wij_Bayesian_zeroerr_eps}; cf. \cite{Knoblauch:NeurComp2011,LansnerRavichandranKnoblauchHerman:arxiv:2025}).
BOMs basically is the generally Bayesian-optimal local learning rule (making, however, the naive-Bayes assumption of independent inputs) and contains
the more commonly used covariance rule \cite{Sejnowski:1977a,Willshaw/Dayan:1990}, BCPNN rule \cite{Lansner/Ekeberg:1989,Lansner:2009,LansnerRavichandranKnoblauchHerman:arxiv:2025}
and (inihibitory) Willshaw/Steinbuch rule \cite{Steinbuch:1961,Willshaw/Buneman/Longuet-Higgins:1969,Palm:1980,Knoblauch/Palm/Sommer:NeurComp2010,Knoblauch:patentIAM:EPOUSJP2012,Knoblauch:2007_b} as
limit cases, depending on activity sparseness \cite{Knoblauch:NeurComp2011}.
As the literal BOMs learning rule (\ref{eq:bj_Bayesian_zeroerr_eps},\ref{eq:wij_Bayesian_zeroerr_eps}) may also be questionable with respect to biological realism,
future experiments should replace BOMs in DCAN layers
by a composition of cortex-like layers including excitatory and inhibitory neurons obeying Dale's law and the simpler learning rules. It may also be interesting to
perhaps combine Oja-like feedwordword PCA-learning with the DCAN model, for example, by using feedforward PCA-learning in the early layers and DCAN with counterstream learning
in the downstream layers. Another interesting variant of DCAN would be to prime counterstream learning by previous feed-forward activity: This would avoid the current inversion
problem, that counterstreams originating from a single one-hot unit coding the correct class cannot really activate the correct upstream cell assemblies that correspond to
the current input being one particular instance of that class. The current implementation of DCAN tries to compensate this inversion problem by using multiple output areas
providing the same targets at multiple (or here even all) hidden layers. 
In a more realistic setting, each (or at least several) layer would receive their own targets (but unlike target propagation from ``external'' sources corresponding to the different
target layers), e.g., from other modalities or within the visual hierachy via horizontal connections (e.g., the WHAT path predicting targets related to the WHERE or HOW paths). 


\appendix

\newpage
\section{Bayesian-optimal model of associative memory (BOM)}\label{app:BOM}
The BOM learning rule has been suggested by \cite{Knoblauch:NeurComp2011} (see also \cite{Knoblauch_BayesAsso:HRI2009,Knoblauch_BayesAssoIJCNN:2010,Knoblauch:AutoBOM:2024,LansnerRavichandranKnoblauchHerman:arxiv:2025}) for storing $M$ memory associations $u^\mu\rightarrow v^\mu$ between binary input patterns $u^\mu\in\{0,1\}^m$
and output patterns $v^\mu\in\{0,1\}^n$ in two neuron layers $u$ and $v$ of size $m$ and $n$, where $\mu=1,\ldots,M$. It uses first order (neural) and second order (synaptic) counter variables
\begin{eqnarray}
  M'_u(i) & := & \#\{\mu: u^{\mu}_i = 1\}             \label{eq:Muprime} \\
  M_v(j) & := & \#\{\mu: v^{\mu}_j = 1\}             \label{eq:Mv} \\
  M_{uv}(i,j) & := & \#\{\mu: u^{\mu}_i = 1, u^{\mu}_j = 1\}      \label{eq:Muv} 
\end{eqnarray}
where $u,v\in\{0,1\}$, $i=1,\ldots,m$ and $j=1,\ldots,n$.
It is actually sufficient to memorize memory number $M$, unit usages $M_1'(i)$, $M_1(j)$, and coincidence counters $M_{11}(i,j)$, as all other counters can be reconstructed
using
\begin{eqnarray}
  M'_0(i) & = & M-M'_1(i)  \label{eq:M0prime} \\
  M_0 (j) & = & M-M_1(j),  \label{eq:M0} \\
  M_{01} (i,j) & = & M_1(j)-M_{11}(i,j)  \label{eq:M01} \\
  M_{10} (i,j) & = & M_1'(i)-M_{11}(i,j)  \label{eq:M10} \\
  M_{00} (i,j) & = & M_0(j)-M_{10}(i,j)=M-M_1(j)-M_1'(i)+M_{11}(i,j)\ .  \label{eq:M00} 
\end{eqnarray}
From this we can compute biases $b_j$ and synaptic weights $w_{ij}$ for postsynaptic neuron $v_j$ as
\footnote{
   For a detailed derivation for biases $b_j$ and weights $w_{ij}$ given in the first equations of (\ref{eq:bj_Bayesian},\ref{eq:wij_Bayesian}) see \cite{Knoblauch:NeurComp2011}, eqs. 2.15-2.17.
   Then the second equations in (\ref{eq:bj_Bayesian},\ref{eq:wij_Bayesian}) follow from inserting (\ref{eq:M0}-\ref{eq:M00}), casting the {\bf four relevant terms}
   as {\bf I)} $M_{01}(1-q_{01|1})+M_{11}q_{10|1}\stackrel{(\ref{eq:M01})}{=}(M_1-M_{11})(1-q_{01|1})+M_{11}q_{10|1}=M_1(1-q_{01|1})-M_{11}(1-q_{01|1}-q_{10|1})$,
   {\bf II)} $M_{00}(1-q_{01|0})+M_{10}q_{10|0}\stackrel{(\ref{eq:M10},\ref{eq:M00})}{=}(M-M_1-M_1'+M_{11})(1-q_{01|0})+(M_1'-M_{11})q_{10|0}=(M-M_{1})(1-q_{01|0})-(M_1'-M_{11})(1-q_{01|0}-q_{10|0})$,
   {\bf III)} $M_{11}(1-q_{10|1})+M_{01}q_{01|1}\stackrel{(\ref{eq:M01})}{=}M_{11}(1-q_{10|1})+(M_1-M_{11})q_{01|1}=M_{11}(1-q_{01|1}-q_{10|1})+M_1q_{01|1}$, and
   {\bf IV)} $M_{10}(1-q_{10|0})+M_{00}q_{01|0}\stackrel{(\ref{eq:M10},\ref{eq:M00})}{=}(M_1'-M_{11})(1-q_{10|0})+(M-M_1-M_1'+M_{11})q_{01|0}=(M_1'-M_{11})(1-q_{01|0}-q_{10|0})+(M-M_1)q_{01|0}$. 
}
\begin{eqnarray}
  b_j   & = & (m-1)\log\frac{M_0}{M_1} + \sum_{i=1}^m \log\frac{M_{01}(1-q_{01|1})+M_{11}q_{10|1}}{M_{00}(1-q_{01|0})+M_{10}q_{10|0}}           \label{eq:bj_Bayesian} \\
        & = & (m-1)\log\frac{M-M_1}{M_1} + \sum_{i=1}^m \log\frac{M_{1}(1-q_{01|1})-M_{11}(1-q_{01|1}-q_{10|1})}{(M-M_{1})(1-q_{01|0})-(M_1'-M_{11})(1-q_{01|0}-q_{10|0})}     \nonumber  \\
  w_{ij} & = & \log\frac{(M_{11}(1-q_{10|1})+M_{01}q_{01|1})(M_{00}(1-q_{01|0})+M_{10}q_{10|0})}{(M_{10}(1-q_{10|0})+M_{00}q_{01|0})(M_{01}(1-q_{01|1})+M_{11}q_{10|1})} \label{eq:wij_Bayesian} \\
        & = & \log\frac{(M_{11}(1-q_{01|1}-q_{10|1})+M_{1}q_{01|1})((M-M_{1})(1-q_{01|0})-(M_1'-M_{11})(1-q_{01|0}-q_{10|0}))}{((M_{1}'-M_{11})(1-q_{01|0}-q_{10|0})+(M-M_{1})q_{01|0})(M_{1}(1-q_{01|1})-M_{11}(1-q_{01|1}-q_{10|1}))} \nonumber
\end{eqnarray}
where we skip indices for brevity, writing $M_v:=M_v(j)$ and $M_{uv}:=M_{uv}(i,j)$, and similarly assume query error estimates
\begin{eqnarray}
  q_{01|v} &:=& q_{01|v}(i,j) := \pr[\tilde{u}_i=1|u^\mu_i=0,v^\mu_j=v] \ ,\label{eq:p01v}\\
  q_{10|v} &:=& q_{10|v}(i,j) := \pr[\tilde{u}_i=0|u^\mu_i=1,v^\mu_j=v] \label{eq:p10v} \ ,
\end{eqnarray}
corresponding to the component transition probabilites of a query input component switching from 0 to 1 or from 1 to 0, conditioned on postsynaptic activity $v$.
Here $\tilde{u}\in\{0,1\}^m$ is the query pattern and
the retrieval task is to return the output pattern $v^\mu$ whose associated input pattern $u^\mu\approx\tilde{u}$ is most similar to the query. Assuming independent input components
$u^\mu_i$ (corresponding to a ``{\bf naive Bayes}'' approach), the optimal (maximum a-posteriori) decision for the retrieval output $\hat{v}\approx v^\mu$ is by choosing components
\begin{eqnarray}
  \hat{v}_j=1 \quad\mbox{if}\quad a_j:=b_j+\sum_{i=1}^nw_{ij}\tilde{u}_i\ \ge 0\quad\mbox{(and $0$ otherwise)}   \label{eq:aj_dendrpotential}
\end{eqnarray}
where the ``dendritic potentials'' $a_j=\ln\frac{\pr[v_j=1|\tilde{u}]}{\pr[v_j=0|\tilde{u}]}$ of output neurons $j$ correspond to the log-odds-ratios.
Note that both nominators and denominators in the log-terms of bias and weights (\ref{eq:bj_Bayesian},\ref{eq:wij_Bayesian}) may become zero, and thus the log-terms {\bf infinity}. To avoid this one may safeguard each nominator and denominator term by a $\max(\epsilon,\ldots)$ operation to {\bf grant numerical stability}.
\footnote{\label{footnote:safeguardingBOM_each_factor}
   It is recommended to {\bf safeguard each factor} in the nominators and denominators of the logarithms in (\ref{eq:bj_Bayesian},\ref{eq:wij_Bayesian}) by a separate $\max(\epsilon,\ldots)$ operation for the sake of
   {\bf numerical stability}.
   The reason is that positive and negative infinite terms in the bias and weights may cancel each other.
   So safeguarding only nominators and denominators as a whole, although possible, may cause problems, where non-zero factors
   will unnecessarily vanish after canceling two infinite factors.
   At least at high memory load only the $M_{11}$ counters may occasionally become zero, and there should be no
   canceling of infinite factors. However, for small networks and/or low memory load also the unit usages $M_1$ and $M_1'$ may sometimes be zero and
   cause the described {\bf cancellation problems}.
   For an {\bf exact implementation} using separate variables for finite and infinite biases and weights see \cite{Knoblauch:NeurComp2011}, appendix A.
   For a {\bf proper choice of the numerical stability parameter} $\epsilon$ in a high memory load regime
   see (\ref{eq:eps_ij_estimates},\ref{eq:eps_estimate_autoasso_Lansner}) below.
}

There exist several {\bf simplified variants of BOM}:
\begin{itemize}\tightitems
\item For {\bf auto-association} the two layers $u=v$ are identical and, correspondingly, the network is recurrent with $M_1'(i)=M_1(i)$ and symmetric
      \begin{align}
        M_{11}(i,j)=M_{11}(j,i) \quad\left(\mbox{or more general $M_{uv}(i,j)=M_{vu}(j,i)$ for $u,v\in\{0,1\}$}\right) .
      \end{align}
      For {\bf autapses} $i=j$ it is obviously
      \begin{align}
         M_{uu}=M_u, \ \  M_{uv}=0\ \mbox{for $u\neq v$}, \ \ q_{01|1}=q_{10|0}=0, \quad \mbox{and $w_{ii}=\log\frac{(1-q_{10|1})\cdot(1-q_{01|0})}{q_{01|0}\cdot q_{10|1}}$.}
      \end{align}
\item For {\bf incompletely connected networks} the sums in (\ref{eq:bj_Bayesian},\ref{eq:aj_dendrpotential}) runs only over the connected presynaptic neurons $i$, where
      $m=m(j)$ corresponds to the number of connections to neuron $j$.
\item {\bf Error estimates} (\ref{eq:p01v},\ref{eq:p10v}) are often represented {\bf without conditioning on postsynaptic activity}, simplifying to
      $q_{01}(i):=\pr[\tilde{u}_i=1|u^\mu_i=0]$ and $q_{10}(i):=\pr[\tilde{u}_i=0|u^\mu_i=1]$ or even to $q_{01}(i):=q_{01}$ and $q_{10}(i):=q_{10}$
      using the same estimates $q_{01}$ and $q_{10}$ for all input neurons $i$.
\item {\bf Iterative retrieval} improves outputs over time. For example, in auto-associative networks, the retrieval output $\hat{u}(\tau)$ at retrieval step $\tau=1,2,\ldots$
      is fedback to the input-layer in the next retrieval step, $\tilde{u}(\tau+1)=\hat{u}(\tau)$. Similarly, iterative retrieval can also be implemented in hetero-associative
      networks if layers $u$ and $v$ are bidirectionally connected \cite{Kosko:1988,Sommer/Palm:1999}.
      Instead of decreasing error estimates (\ref{eq:p01v},\ref{eq:p10v}) with retrieval steps
      (which requires a form of {\bf short-term plasticity} like synaptic faciliation or depression \cite{Knoblauch:AutoBOM:2024,Zucker/Regehr:2002,Hennig:2013}), a further simplification
      may employ {\bf fixed error estimates},
      corresponding either to {\bf initial error levels} in the original query $\tilde{u}(1)$ or to {\bf final (low or zero) error levels} in the retrieval outputs $\hat{v}(\tau)$
      for $\tau\rightarrow\infty$. Preliminary experiments suggest that the latter strategy is most
      beneficial \cite{Knoblauch:AutoBOM:2024,LansnerRavichandranKnoblauchHerman:arxiv:2025} and simplifies the learning procedure significantly, as is shown in the following.
\end{itemize}
For {\bf zero error estimates} $q_{01|v}=q_{10|v}=0$, bias and synaptic weights (\ref{eq:bj_Bayesian},\ref{eq:wij_Bayesian}) simplify to  
\begin{eqnarray}
  b_j   & = & (m-1)\log\frac{M_0}{M_1} + \sum_{i=1}^m \log\frac{M_{01}}{M_{00}} \ =\  (m-1)\log\frac{1-p_j}{p_j} + \sum_{i=1}^m \log\frac{p_j-p_{ij}}{1-p_i'-p_j+p_{ij}} \quad\quad\quad  \label{eq:bj_Bayesian_zeroerr} \\
  w_{ij} & = & \log\frac{M_{11}M_{00}}{M_{10}M_{01}}\ =\ \log\frac{p_{ij}\cdot (1-p_i'-p_j+p_{ij})}{(p_i'-p_{ij})\cdot (p_j-p_{ij})}  \label{eq:wij_Bayesian_zeroerr}
\end{eqnarray}
where the latter equations cast the BOM learning rule
\footnote{
   We often call (\ref{eq:bj_Bayesian_zeroerr},\ref{eq:wij_Bayesian_zeroerr}) (and (\ref{eq:bj_Bayesian_zeroerr_eps},\ref{eq:wij_Bayesian_zeroerr_eps})) the {\bf simplified} or {\bf symmetric BOM} rule ({\bf BOMs}) because in the case of auto-association it produces
   a {\bf symmetric weight matrix} \cite{Knoblauch:AutoBOM:2024,LansnerRavichandranKnoblauchHerman:arxiv:2025}.
   Of course, also the {\bf full BOM} rule (\ref{eq:bj_Bayesian},\ref{eq:wij_Bayesian}) can be written using the counter fractions (\ref{eq:pi_pij_counterfractions}), where we obtain
   $b_j=(m-1)\log\frac{1-p_j}{p_j} + \sum_{i=1}^m \log\frac{p_j(1-q_{01|1})-p_{ij}(1-q_{01|1}-q_{10|1})}{(1-p_j)(1-q_{01|0})-(p'_i-p_{ij})(1-q_{01|0}-q_{10|0})}$ and
   $w_{ij}=\log\frac{(p_{ij}(1-q_{01|1}-q_{10|1})+p_jq_{01|1})((1-p_j)(1-q_{01|0})-(p'_i-p_{ij})(1-q_{01|0}-q_{10|0}))}{((p'_i-p_{ij})(1-q_{01|0}-q_{10|0})+(1-p_j)q_{01|0})(p_j(1-q_{01|1})-p_{ij}(1-q_{01|1}-q_{10|1}))}$.
}
in the notation of \cite{Minai:1997,LansnerRavichandranKnoblauchHerman:arxiv:2025}, wherefore
we expand the counters $M'_u$, $M_v$, $M_{uv}$ by factor $M$, insert (\ref{eq:M0}-\ref{eq:M00}), and define corresponding
{\bf counter fractions}
\begin{eqnarray}
   p_i':=\frac{M_1'(i)}{M}, \quad p_j:=\frac{M_1(j)}{M}, \quad p_{ij}:=\frac{M_{11}(ij)}{M}\quad\in[0;1]\ .   \label{eq:pi_pij_counterfractions}
\end{eqnarray}
To avoid division by $0$ and $\log(0)$ we may bound each factor in nominators and denominators by a small {\bf numerical stabilization parameter} $\eps>0$ (see also footnote~\ref{footnote:safeguardingBOM_each_factor}),
\begin{eqnarray}
  b_j   & = & (m-1)\log\frac{\max(1-p_j,\eps)}{\max(p_j,\eps)} + \sum_{i=1}^m \log\frac{\max(p_j-p_{ij},\eps)}{\max(1-p_i'-p_j+p_{ij},\eps)}   \label{eq:bj_Bayesian_zeroerr_eps} \\
  w_{ij} & = & \log\frac{\max(p_{ij},\eps)\cdot \max(1-p_i'-p_j+p_{ij},\eps)}{\max(p_i'-p_{ij},\eps)\cdot \max(p_j-p_{ij},\eps)}  \label{eq:wij_Bayesian_zeroerr_eps}
\end{eqnarray}
{\bf In practice} we can assume high memory load close to the capacity limit $M\sim\frac{m}{-q(1-q)\log(q)}\gg m,n$ (see \cite{Knoblauch:NeurComp2011}, eq.~2.37) for sparse neural activity
with $p:=p[u^\mu_i=1]\ll\frac{1}{2}$ and $q:=p[v^\mu_j=1]\ll\frac{1}{2}$ \cite{Waydo/Kraskov/Quiroga/Fried/Koch:2006}, where all neurons $i,j$ have positive unit usages $M_1'(i),M_1(j)>0$
(otherwise we or structural plasticity may remove unused neurons \cite{Knoblauch:Elsevier2017,Ming/Song:2005,Holtmaat/Svoboda:2009}). Consequently, we will get 
positive synaptic counters $M_{00},M_{01},M_{10}>0$, and only the {\bf coincidence counters} $M_{11}(i,j)$ may remain {\bf zero sometimes},
\footnote{
  For example, assuming {\bf random patterns} with independent components, all {\bf synaptic counters} have {\bf binomial distributions} with {\bf expectations}
  $E(M_{11})=Mpq$, $E(M_{10})=Mp(1-q)$, $E(M_{01})=M(1-p)q$, $E(M_{00})=M(1-p)(1-q)$ and {\bf variances} $\mathrm{Var}(M_{11})=Mpq(1-pq)$, $\mathrm{Var}(M_{10})=Mp(1-q)(1-p(1-q))$,
  $\mathrm{Var}(M_{01})=M(1-p)q(1-(1-p)q)$, $\mathrm{Var}(M_{00})=M(1-p)(1-q)(1-(1-p)(1-q))$. Inserting the {\bf capacity limit} $M\sim \frac{m}{-q(1-q)\log(q)}$
  of large networks with $m\rightarrow\infty$ and sparse activity $p,q\rightarrow 0$ (see \cite{Knoblauch:NeurComp2011}, eq.~2.37), we get
  for the ``mixed counters'' quickly diverging expectations $E(M_{01})\sim\frac{m}{-\log q}\rightarrow\infty$ and $E(M_{10})\sim\frac{mp}{-q\log q}\rightarrow\infty$, whereas
  $E(M_{11})\sim\frac{mp}{-\log q}$ may remain constant or diverges only slowly, dependent on sparsity level of activity. Since obviously $E(M_{uv})\sim \mathrm{Var}(M_{uv})$ for $uv\in\{11,10,01\}$
  we obtain also {\bf quickly diverging expectation-to-standard-deviation ratios} $\frac{E(M_{uv})}{\sqrt{\mathrm{Var}(M_{uv})}}\sim\sqrt{E(M_{uv})}\rightarrow\infty$
  for $M_{10}$ and $M_{01}$, but not for $M_{11}$.
  As a consequence (applying Chebyshev's inequality), we will have $M_{10},M_{01}\gg 0$ with very high probability, whereas $M_{11}$ may sometimes remain zero.
  As for $M_{00}$ with $p,q\rightarrow 0$ the chance of remaining zero
  is obviously much smaller than for $M_{10},M_{01}$, we have also $M_{00}\gg 0$ with very high probability.  
  Thus, {\bf we have shown} that all counter variables except $M_{11}$ (and as a consequence, also the unit usages $M_1\ge M_{01}$ and $M_1'\ge M_{10}$) will be positive with very high probability,
  and {\bf only the coincidence counters} $M_{11}$ may sometimes {\bf remain zero}.  
}
causing numerical problems due to
strongly inhibitory synapses \cite{Knoblauch:2007_b,Knoblauch:NeurComp2011} with $w_{ij}\rightarrow-\infty$ for $\epsilon\rightarrow 0$.
Thus, to find a {\bf reasonable value} for $\eps$, we may identify $\eps\cdot M$ with the relevant error term $M_{01}q_{01|1}$ in the left bracket of the nominator in (\ref{eq:wij_Bayesian})
and choose
\begin{eqnarray}
   \eps(i,j) := \frac{M_{01}(i,j)\cdot q_{01|1}(i,j)}{M}\approx \frac{M_{01}(i,j)\cdot q_{01}(i)}{M} \approx (1-p)\cdot q\cdot q_{01} \ =:\ \eps   \label{eq:eps_ij_estimates}
\end{eqnarray}
where we approximated $M_{01}\approx M(1-p)q\approx Mq$ by its mean value assuming a binomial distribution $M_{01}(i,j)\sim B(M,(1-p)\cdot q)$.
Note that (\ref{eq:eps_ij_estimates}) provides {\bf synapse-specific stabilization parameters} $\eps(i,j)$ for the critical max-operation in the nominator of (\ref{eq:wij_Bayesian_zeroerr_eps})
that can easily be computed from the synaptic counter $M_{01}(i,j)$
and query error estimates $q_{01|1}(i,j)$ or $q_{01}(i)$. By contrast, the last approxation in (\ref{eq:eps_ij_estimates})
corresponds to a {\bf single general estimate} $\eps$ that is independent of indexes $i,j$ and may be applied either only for
the remaining non-critical (unlikely) max-operations in (\ref{eq:bj_Bayesian_zeroerr_eps},\ref{eq:wij_Bayesian_zeroerr_eps}) or for all max-operations to
enforce a symmetric weight matrix $w_{ij}=w_{ji}$ for auto-association \cite{LansnerRavichandranKnoblauchHerman:arxiv:2025}.
For example, for (either auto-associative or bidirectional) {\bf iterative retrieval} with {\bf $k$-winners-take-all activation}
and mean activity fraction $p=k/m$ in (input) layer $u$, 
targeting at a {\bf minimum fraction} $p_{\mathrm{corr}}$ of {\bf correct retrieval outputs} $\hat{u}=u^\mu$, we may use
\footnote{
   Due to $k$-winners-take-all retrieval in input layer $u$ (assuming either auto-associative or bidirectional iterative retrieval),
   a correct retrieval result $\hat{u}$ in the input layer
   means that none of the $m-k$ silent neurons in $u^\mu$ have switched to active neurons.
   Thus, the probability of a correct retrieval result
   is $p_{\mathrm{corr}}\approx(1-q_{01})^{m-k}$ assuming again a binomial distribution of errors (and $q_{01}$ being the error probability
   in the {\it late} phase of iterative retrieval). Solving for $q_{01}$ gives $q_{01}\approx 1-p_{\mathrm{corr}}^{1/(m-k)}$.
   Alternatively, taking logarithms gives
   for small error probabilities $\log(p_{\mathrm{corr}})\approx (m-k)\log(1-q_{01})\approx -(m-k)q_{01}=-m(1-p)q_{01}$ or $q_{01}\approx \frac{-\log(p_{\mathrm{corr}})}{m(1-p)}$.
   Inserting this in (\ref{eq:eps_ij_estimates}) gives (\ref{eq:eps_estimate_autoasso_Lansner}).
}
\begin{eqnarray}
   q_{01}\approx 1-p_{\mathrm{corr}}^{1/(m-k)}\approx\frac{-\log(p_{\mathrm{corr}})}{m(1-p)} \quad\mbox{or}\quad \eps \approx  (1-p)\cdot q \cdot (1-p_{\mathrm{corr}}^{1/(m-k)}) \approx \frac{-q\cdot \log(p_{\mathrm{corr}})}{m}  \ . \quad \label{eq:eps_estimate_autoasso_Lansner}
\end{eqnarray}
With a similar reasoning, we may use error probabilities $q_{01}$ as in (\ref{eq:eps_estimate_autoasso_Lansner})
and $q_{10}=q_{01}\cdot\frac{m-k}{k}\approx \frac{-\log(p_{\mathrm{corr}})}{k}$ also for $k$-winners-take-all iterative retrieval with the
{\bf asymmetric BOM rule} (\ref{eq:bj_Bayesian},\ref{eq:wij_Bayesian}) {\bf without conditioning}, 
\footnote{
  The $k$-winners-take-all selection in layer $u$ grants $k$ active neurons in each pattern. Therefore the average
  number of false positive bit errors $f_{01}:=(m-k)q_{01}$ must equal the corresponding false negative errors $f_{10}:=kq_{10}$.
  Hence $(m-k)q_{01}=kq_{10}$ or $q_{10}=q_{01}\cdot\frac{m-k}{k}$.
}
which in practice is often the {\bf best model} unless requiring symmetric weights.

\newpage
\section{Exponentially weighted moving average modulated by dopamine (DAMA)}\label{app:EWMA}
For continuous learning we want to avoid neuronal and synaptic counters approaching infinity. Therefore, we apply for time dependent
counter variables like $M(\tau)$ the recursion
\begin{align}
   \tilde{M}(\tau) = \beta\tilde{M}(\tau-1)+(1-\beta)M(\tau)  \quad\mbox{for $\beta\in[0;1)$, $\tau=1,2,\ldots$}  \label{eq:tildeM_recursion}
\end{align}
for realizing an (uncorrected) {\bf exponentially weighted moving average} \cite{Kingma/Ba:2015}. 
Specifically, expanding the recursion for {\bf initial value} $\tilde{M}(0):=0$ yields
\begin{align}
  \tilde{M}(\tau) &= (1-\beta)M(\tau) + (1-\beta)\beta M(\tau-1) + (1-\beta)\beta^2 M(\tau-2) + \ldots + (1-\beta)\beta^{\tau-1}M(1) \nonumber\\
                  &= \sum_{t=0}^{\tau-1}(1-\beta)\beta^tM(\tau-t)  = \sum_{t=1}^\tau(1-\beta)\beta^{\tau-t}M(t)   \label{eq:tildeM_recursion_expansion}
\end{align}
which is a weighted sum of the counter values $M(t)$ for $t=0,1,2,\ldots,\tau$, corresponding to a discrete convolution.
Using the geometric sum, the weights $(1-\beta)\beta^{\tau-t}$ of $M(t)$ sum up to
\begin{align}
  \sum_{t=1}^\tau(1-\beta)\beta^{\tau-t} = (1-\beta)\sum_{t=0}^{\tau-1}\beta^{t}=1-\beta^{\tau}\ .  \label{eq:tildeM_sum_of_weights}
\end{align}
As for a moving average the sum of weights should be 1, we define the correction
\footnote{
   If all $M(\tau)$ have the same mean value $\mu:=E(M(\tau))$, a consistent moving average $\bar{M}(\tau)$ should have
   the identical mean $E(\bar{M}(\tau))\stackrel{!}{=}\mu$. However, for (\ref{eq:tildeM_recursion_expansion}) we would
   get $E(\tilde{M}(\tau))=\sum_{t=1}^\tau(1-\beta)\beta^{\tau-t}E(M(t))=(1-\beta^{\tau})\mu$ from the linearity of expectatation.
   Therefore we correct by (\ref{eq:barM_tau}) and get $E(\bar{M}(\tau))=\frac{E(\tilde{M}(\tau))}{1-\beta^{\tau}}=\mu$ as desired. 
}
\begin{align}
  \bar{M}(\tau) := \frac{\tilde{M}(\tau)}{1-\beta^{\tau}} \label{eq:barM_tau}
\end{align}
which is called the {\bf corrected exponentially weighted moving average}.
Note that the correction terms become negligible with time, $\bar{M}(\tau)\rightarrow \tilde{M}(\tau)$ for $\tau\rightarrow\infty$.

Many reinforcement-learning algorithms modulate learning by the temporal-difference-learning error \cite{Sutton/Barto:1998},
which correlates with phasic {\bf dopamine release} $\delta(\tau)\ge 0$ \cite{Schultz/Dayan/Montague:1997}.
To account for this in the recursion (\ref{eq:tildeM_recursion}), we scale the contribution $(1-\beta)M(\tau)$ of counter $M(\tau)$ at time $\tau$
by factor $\delta(\tau)$ and obtain the {\bf DopAmine-modulated Moving Average (DAMA)}
\begin{align}
   \tilde{M}(\tau)    &= \beta\tilde{M}(\tau-1)+(1-\beta)\delta(\tau)M(\tau)  \quad\mbox{for $\beta\in[0;1)$, $\tau=1,2,\ldots$}          \nonumber \\
   \bar{\delta}(\tau) &= \beta\bar{\delta}(\tau-1)+(1-\beta)\delta(\tau) \quad\left(= \sum_{t=1}^\tau(1-\beta)\delta(t)\beta^{\tau-t}\right) \nonumber \\
   \bar{M}(\tau)      &= \frac{\tilde{M}(\tau)}{\bar{\delta}(\tau)}                  \label{eq:DAMA}
\end{align}
where, similar as before, $\tilde{M}(\tau)$ is the uncorrected DAMA, $\bar{\delta}(\tau)$ is a moving average of dopamine, corresponding to the sum of weights
for DAMA, and $\bar{M}(\tau)$ is the corrected DAMA of counter variable $M(\tau)$.

\subsubsection*{Acknowledgments} I am grateful to Anders Lansner and Patrick Inoue for valuable discussions.

\newpage

%
%
\setlength{\itemsep}{0cm}
\bibliographystyle{plain}
\bibliography{../../../../../03-resources/LATEX_gen/neuroAK}

\end{document}